\documentclass[lettersize,journal]{IEEEtran}
\usepackage{amsmath,amsfonts}
\usepackage{algorithmic}
\usepackage{algorithm}
\usepackage{array}
\usepackage[caption=false,font=normalsize,labelfont=sf,textfont=sf]{subfig}
\usepackage{textcomp}
\usepackage{stfloats}
\usepackage{url}
\usepackage{verbatim}
\usepackage{graphicx}
\usepackage{orcidlink}
\usepackage[square,numbers,sort&compress]{natbib}
\usepackage{tabularx}
\usepackage{xcolor}
\usepackage{array}
\usepackage{tabularx}
\usepackage{makecell}
\usepackage{array}
\usepackage{tabularx}
\usepackage{array}
\usepackage{enumitem}
\usepackage{multirow}
\usepackage{booktabs}

\newcolumntype{Y}{>{\raggedright\arraybackslash}X}
\begin{document}

\title{{\fontsize{17}{15}\selectfont Decoding the Disaster: Multi-Task Geospatial Reasoning with Vision-Language Models and Crowdsourced Imagery for Disaster Mapping}}


\author{
Wenping Yin\textsuperscript{\orcidlink{0009-0002-2056-0091}},
Fabian Deuser\textsuperscript{\orcidlink{0000-0003-4511-4223}},
Ziqi Liu\textsuperscript{\orcidlink{0009-0001-0483-5502}},
Naixia Mou\textsuperscript{\orcidlink{0000-0003-1700-5943}},
Weijia Li\textsuperscript{\orcidlink{0000-0003-1838-9176}},~\IEEEmembership{Member,~IEEE},\\
Pedram Ghamisi \textsuperscript{\orcidlink{0000-0003-1203-741X}},~\IEEEmembership{Senior Member,~IEEE}, 
Xiao Xiang Zhu\textsuperscript{\orcidlink{0000-0001-5530-3613}},~\IEEEmembership{Fellow,~IEEE},
Hao Li\textsuperscript{\orcidlink{0000-0002-6336-8772}},~\IEEEmembership{Member,~IEEE},

\thanks{This work was supported by the Start-Up Grant (SUG) project “Geospatial Artificial Intelligence for Climate Resilient Urban Environment” from the National University of Singapore (E-109-00-0036-01), and the MoE Tier1 project "Assessing Urban Flood Resilience against Climate Extreme with GeoAI in Southeast Asia". \textit{(Corresponding author: Hao Li)}}
\thanks{Wenping Yin is with the College of Geodesy and Geomatics, Shandong University of Science and Technology, Qingdao 266590, China (email: wenpingyin2019@gmail.com).}
\thanks{Fabian Deuser is with the Professorship of Big Geospatial Data Management, School of Engineering and Design, Technical University of Munich, Munich, 85521, Bavaria, Germany (email: fabian.deuser@tum.de).}
\thanks{Ziqi Liu is with the State Key Laboratory of Information Engineering in Surveying, Mapping and Remote Sensing, Wuhan University, Wuhan 430079, China (email: lzq677@whu.edu.cn).}
\thanks{Naixia Mou is with the College of Geodesy and Geomatics, Shandong University of Science and Technology, Qingdao 266590, China (email: mounaixia@163.com).}
\thanks{Weijia Li is with the Tsinghua Shenzhen International Graduate School, Tsinghua University, Shenzhen 518055, China. (email: liweijia@sz.tsinghua.edu.cn).} 
\thanks{Pedram Ghamisi is with Helmholtz-Zentrum Dresden-Rossendorf, Responsible AI Group, 09599 Freiberg, Germany and also with the University of Iceland, Faculty of Electrical and Computer Engineering, Reykjavik 102, Iceland (e-mail: p.ghamisi@hzdr.de).}
\thanks{Xiao Xiang Zhu is with the Chair of Data Science in Earth Observation, Technical University of Munich, 80333 Munich, Germany and with the Munich Center for Machine Learning, 80333 Munich, Germany (email: xiaoxiang.zhu@tum.de).}
\thanks{Hao Li is with the Department of Geography, National University of Singapore, Singapore 117568, Singapore (email: hao.li@nus.edu.sg).}

}

\markboth{Submitted to IEEE Geoscience and Remote Sensing Magazine}%
{Yin \MakeLowercase{\textit{et al.}}: Decoding the Disaster}


\maketitle
\begin{abstract}
Crowdsourced imagery provides timely, fine-grained, street-level observations for disaster mapping, complementing conventional remote sensing imagery (RSI) during emergency response. However, such imagery is often unstructured, spatially ambiguous, and lacks reliable geographic metadata, making manual geolocalization and interpretation labor-intensive and difficult to scale. This work proposes a multi-task \textbf{G}eospatial \textbf{R}easoning \textbf{Disaster} mapping framework, namely \textbf{GRDisaster}, to examine the potential of vision–language models (VLMs) in understanding, geolocalizing, and reasoning over crowdsourced disaster imagery. \textbf{GRDisaster} is built on a newly curated benchmark dataset derived from PhotoMappers, comprising 26,340 images organized into human-validated volunteered geographic information (VGI)--street-view imagery (SVI)--RSI cross-view triplets covering multiple disaster events from 2018 to 2024. The framework combines deterministic and probabilistic cross-view geolocalization with multi-view fusion to associate VGI images with georeferenced SVI and RSI. It introduces two sets of spatial reasoning indicators for cross-view geolocalization validation and disaster damage assessment. These indicators use structural, environmental, and global-scene cues to validate cross-view correspondences and visually observable damage evidence with expert-verified annotations to assess disaster severity, improving the interpretability of VLM outputs. To our knowledge, this study provides the first systematic investigation and unified evaluation framework for examining how VLM-based spatial reasoning can transform crowdsourced disaster imagery into actionable geospatial artificial intelligence (GeoAI) through cross-view geolocalization validation, interpretable spatial reasoning, and damage-aware severity assessment. The findings provide timely guidance for developing scalable, interpretable, and responsible GeoAI-assisted disaster mapping systems.

\end{abstract}


\section{Introduction}
\IEEEPARstart{T}{imely} and accurate disaster mapping is essential for emergency response, damage assessment, and resource allocation \citep{li2025cross, zhao2026causal, shahi2026governing}. During major disasters, crowdsourced imagery such as volunteered geographic information (VGI) provides rich street-level observations that complement remote sensing imagery (RSI), particularly when satellite observations are delayed, obstructed by clouds, or insufficient to capture fine-grained local damage \citep{zhu2023geoinformation, yin2025llm, yin2026triple}. However, these images are often unstructured, noisy, spatially ambiguous, and lacking reliable geographic metadata, making manual geolocalization and interpretation labor-intensive and difficult to scale \citep{deuser2023sample4geo, saralioglu2020crowdsourcing}. This creates a major bottleneck in disaster response, where timely geolocalization, interpretation, and validation of visual evidence are critical for transforming scattered observations into actionable geospatial artificial intelligence (GeoAI).

Effective disaster response requires georeferenced situational awareness and timely damage assessment, but crowdsourced imagery collected during early response windows often lacks reliable coordinates due to degraded global navigation satellite system signals, missing metadata, or inaccurate user-provided geolocations \citep{huang2025sar, zheng2026weakly, xing2024synergy}. Although satellite and aerial imagery provides broad spatial coverage, it often cannot capture fine-grained street-level damage to buildings, roads, infrastructure, and vegetation. Cross-view geolocalization helps bridge this gap by anchoring ground-level VGI observations to georeferenced SVI and RSI references, recovering coordinates needed for mapping, routing, and operational decision-making \citep{ye2025cross, li2025unsupervised}. However, disaster scenes are difficult to match because post-event imagery may differ substantially from reference data due to collapse, flooding, debris, vegetation loss, and road blockage. Reliable analysis must go beyond appearance similarity and reason over persistent spatial cues, damage indicators, and changes in road topology, building layout, landmarks, and vegetation patterns.

Vision--language models (VLMs) provide a promising pathway to address these challenges in disaster mapping \citep{xiong2024earthnets, weng2025vision, zhu2026foundations}. High-quality post-event annotations are often scarce, delayed, and unevenly distributed across regions and disaster types, limiting the scalability of supervised training. VLMs can interpret heterogeneous visual and textual evidence through open-vocabulary and instruction-guided reasoning, making them suitable for unfamiliar disaster scenes \citep{ghamisi2025responsible, li2026full}. For geolocalization, they can compare candidate matches, identify persistent features, and reason about spatial relationships across ground-level and overhead views \citep{yin2026sagingeo}. For damage assessment, they can recognize affected objects, estimate damage severity, and describe observed changes in their geographic context \citep{wen2026remote, liu2025rescueadi}. Despite these capabilities, the potential of VLMs for interpretable geolocalization and damage assessment in disaster mapping remains largely underexplored.

\begin{figure*}
    \centering
    \includegraphics[width=\linewidth]{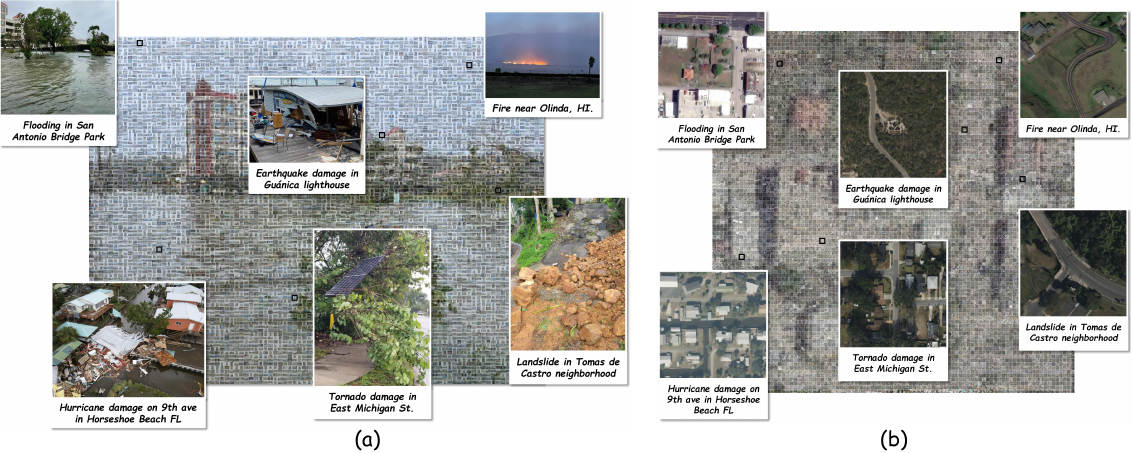}
    \caption{Examples from the PhotoMappers dataset: (a) crowdsourced VGI and (b) corresponding RSI across different disaster types.}
    \label{fig:dataset}
\end{figure*}

This paper proposes \textbf{GRDisaster}, a multitask VLM-based \textbf{G}eospatial \textbf{R}easoning framework for \textbf{Disaster} mapping. GRDisaster is built on a newly curated PhotoMappers dataset comprising human-validated crowdsourced disaster VGI from multiple events between 2018 and 2024. Representative VGI images and corresponding RSI across different disaster types are shown in Fig.~\ref{fig:dataset}. For cross-view geolocalization, GRDisaster matches VGI images with candidate RSI and uses VLM-based reasoning to evaluate visual appearance, spatial layout, and landmark cues, producing geolocalization decisions and explanations. It also classifies damage severity and explains the supporting visual evidence. The major contributions of GRDisaster are summarized as follows:

\begin{itemize}
\item To establish a human-validated VGI--SVI--RSI benchmark dataset covering multiple disaster events for cross-view geolocalization and damage assessment.
\item To formulate cross-view validation as a structured spatial-reasoning task using appearance, spatial-layout, and landmark evidence while providing interpretable rationales.
\item To integrate cross-view geolocalization with explainable VLM-based reasoning for match verification and damage severity classification within a unified workflow.
\end{itemize}

The remainder of this paper is organized as follows. Section 2 reviews related work on GeoAI and multimodal remote sensing for disaster response, VLM-based geospatial reasoning, and crowdsourced multitask disaster mapping. Section 3 presents the design of GRDisaster, including cross-view geolocalization, VLM-based spatial validation of VGI--SVI--RSI correspondences, and VGI-based disaster damage severity assessment. Section 4 presents experiments on the curated PhotoMappers dataset, covering cross-view geolocalization performance and VLM-based spatial reasoning for geolocalization validation and disaster damage severity classification. Section 5 discusses the main findings, limitations, and broader implications for disaster mapping and GeoAI. Section 6 concludes the paper and summarizes the potential of GRDisaster for scalable, trustworthy, and interpretable disaster mapping.

\section{Related work} \label{RelatedWork}
\subsection{GeoAI and multimodal RS in disaster response}

GeoAI has become a key foundation for disaster response by enabling spatially explicit learning, prediction, and decision support from heterogeneous geospatial data \citep{Janowicz2020GeoAI}. Compared with manual mapping or rule-based image analysis, GeoAI methods can extract disaster-related information from RSI, including inundated areas, damaged buildings, blocked roads, burned regions, and disrupted infrastructure. Recent deep learning methods, including convolutional neural networks, object-based semantic change detection, and transformer-based architectures, have improved rapid damage assessment and emergency response. Benchmarks such as xBD have promoted automated building geolocalization and damage classification using pre- and post-disaster satellite imagery \citep{gupta2019creating}. Deep object-based change detection methods further demonstrate the potential of integrating image segmentation, object geolocalization, and semantic damage interpretation in post-disaster scenarios \citep{zheng2021building}. These studies show that GeoAI can transform Earth observation data into actionable disaster information, but single-source overhead imagery remains limited in capturing fine-grained, street-level, and context-dependent damage conditions.

Multimodal RS has become important for disaster response because different sensing modalities provide complementary spatial, spectral, temporal, and semantic information. Optical satellite imagery offers intuitive visual interpretation and high-resolution land-surface details, whereas synthetic aperture radar (SAR) provides observations under cloud cover and adverse weather conditions, which is valuable during floods, hurricanes, and storms. Unmanned aerial vehicle imagery provides flexible and fine-grained local observations, while street-view imagery and crowdsourced VGI capture human-scale disaster impacts that may not be visible from overhead views. Crowdsourcing has been widely discussed as a complement to conventional RS for collecting timely and locally detailed geospatial observations \citep{saralioglu2020crowdsourcing, xu2026towards}. Multi-source multimodal disaster-response studies have also shown that combining satellite, aerial, ground-level, and social media data can improve situational awareness across preparedness, response, recovery, and resilience planning \citep{li2025learning, yin2026disastertd}. Social media-based geoinformation harvesting further highlights the role of community RS in complementing traditional Earth observation systems \citep{zhu2023geoinformation}. However, multimodal integration remains challenging because disaster data are often noisy, incomplete, temporally misaligned, and unevenly distributed across geolocations and events. Crowdsourced images may contain rich local evidence but often lack reliable geographic metadata, making them difficult to integrate directly into operational disaster maps.

Recent GeoAI-assisted disaster mapping studies have moved from single-task image interpretation toward integrated frameworks that combine geolocation recovery, cross-view matching, damage severity assessment, and scene understanding. This direction is particularly relevant for multimodal disaster response, where overhead RSI, street-view references, and crowdsourced ground-level observations need to be interpreted jointly. Cross-view geolocalization methods can anchor unstructured disaster photos to georeferenced RS and street-view references \citep{deuser2023sample4geo,ye2025satellite,tang2025r2ploc,li2025cross}. Language and multimodal models have also been used to exploit implicit geoinformation from disaster-related multimodal data \citep{ye2025cross,yin2026triple}. By linking visual evidence with semantic and spatial cues, these models can support interpretable analysis of buildings, roads, vegetation, landmarks, and disaster-related damage patterns.

\subsection{VLMs and geospatial reasoning}
VLMs have recently become an important direction for connecting visual perception with natural language understanding \citep{lin2025rs, li2025knowledge}. Early models such as CLIP learn transferable visual representations by aligning images and text at large scale, enabling zero-shot recognition and image--text retrieval without task-specific labels \citep{radford2021learning}. Later multimodal models further extended this idea from representation learning to instruction-following and visual dialogue. For example, Flamingo introduced few-shot visual language learning across interleaved image-text inputs \citep{alayrac2022flamingo}, while BLIP-2 connected frozen image encoders with large language models through a lightweight querying transformer. These models show strong potential for open-vocabulary visual interpretation, but their original design is mainly based on natural images and general web data, which makes direct transfer to geospatial applications non-trivial.

Remote sensing and geospatial scenes introduce additional challenges for VLMs \citep{hong2026foundation}. Unlike natural images, RS imagery often contains large spatial extents, small and densely distributed objects, multi-scale patterns, nadir-view geometry, and domain-specific land-cover semantics. A model must not only recognize objects, but also reason about spatial layout, relative position, scene function, and cross-view consistency. To address this gap, recent studies have developed remote-sensing-specific VLMs and datasets. RemoteCLIP adapts vision--language pretraining to RS imagery and improves zero-shot classification, retrieval, and object counting in remote sensing scenarios \citep{liu2024remoteclip}. RS5M and GeoRSCLIP further construct a large-scale RS image-text dataset and domain-pretrained vision--language model for remote sensing tasks \citep{zhang2024rs5m}. Recent work such as DOFA-CLIP further explores multimodal vision--language foundation models for Earth observation by aligning heterogeneous EO modalities with language \citep{xiong2025dofa}. These efforts indicate that domain-specific image-text alignment is important for transferring VLMs from generic visual recognition to geospatial understanding.

More recent work has begun to move from image-level recognition toward geospatial reasoning and grounded interpretation. GeoChat, for example, supports conversational understanding, region-level reasoning, visual grounding, and referring tasks for high-resolution RS images \citep{kuckreja2024geochat}. EarthGPT further explores multimodal large language models for multi-sensor RS image comprehension, including optical, SAR, and infrared imagery \citep{zhang2024earthgpt}. TerraScope extends this direction toward pixel-grounded visual reasoning for Earth observation, emphasizing fine-grained spatial evidence in geospatial reasoning tasks \citep{shu2026terrascope}. These models suggest that VLMs can support more flexible geospatial interpretation by linking visual evidence with textual prompts, spatial cues, and task-specific reasoning requirements. However, disaster scenarios remain particularly challenging because post-event imagery often contains occlusion, debris, damage-induced appearance changes, and missing or unreliable geographic metadata.

\subsection{Crowdsourcing and multi-task disaster mapping}

Crowdsourced data complement conventional remote sensing by providing timely, street-level information from disaster-affected areas \citep{cheng2026beyond, ye2025cross}. Social media posts and VGI imagery can reveal local flooding, structural damage, road accessibility, and infrastructure disruption that may be difficult to observe from overhead imagery alone. Crisis-mapping efforts following events such as the Haiti earthquake demonstrated the value of volunteered data when official information is incomplete or delayed \citep{earle2011twitter}. Their operational use, however, remains constrained by uneven coverage, uncertain data quality, duplicate content, weak annotations, and missing or inaccurate geographic metadata. Addressing these limitations is essential for converting crowdsourced imagery into reliable information for disaster preparedness, response, and recovery.

\begin{figure*}
    \centering
    \includegraphics[width=\linewidth]{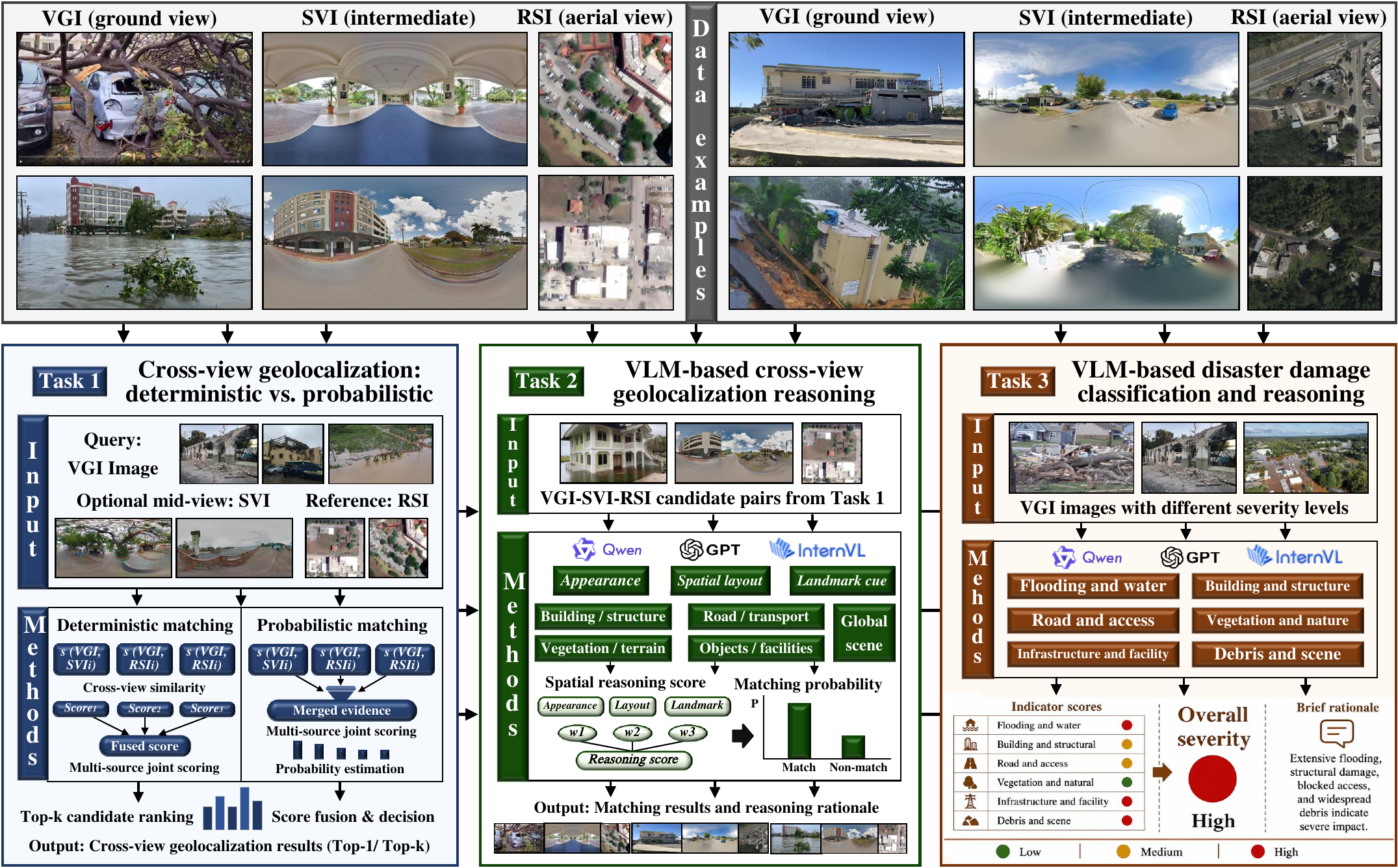}
    \caption{Overview of the three interconnected tasks in the GRDisaster framework using the VGI--SVI--RSI benchmark: deterministic and probabilistic cross-view geolocalization, and explainable VLM-based reasoning for match verification and damage severity classification.}
    \label{fig:overall}
\end{figure*}

Recent disaster mapping research has also moved toward multi-task frameworks that jointly address several operational needs rather than treating each task separately. This trend is consistent with broader Earth observation benchmark efforts, where diverse datasets, tasks, and foundation models are organized for systematic evaluation \citep{xiong2024earthnets}. Recent extreme-event benchmarks further highlight the need to evaluate foundation models across multiple disaster-related tasks, data sources, and event types \citep{zhao2025exebench}. In practical response workflows, geolocation recovery, damage assessment, scene interpretation, and evidence validation are closely connected: an image is more useful when its geolocation is known, its damage level is estimated, and its visual evidence can be explained in relation to surrounding geospatial context. Large-scale datasets such as xBD have supported building geolocalization and damage classification from satellite imagery \citep{gupta2019creating}, while cross-view disaster geolocalization studies have explored how ground-level disaster images can be matched with street-view and remote sensing references \citep{li2025cross,yin2026triple}. For crowdsourced data, this multi-task perspective is particularly important because the same image may simultaneously contain clues for geolocalization, damage severity classification, and scene-level reasoning. Therefore, combining crowdsourced imagery with multi-task disaster mapping can help transform scattered visual observations into more reliable and interpretable geospatial intelligence for emergency response.

\section{Method} \label{Method}

\subsection{Overview}

As shown in Fig. \ref{fig:overall}, GRDisaster consists of three interconnected tasks: cross-view geolocalization, VLM-based cross-view reasoning and match validation, and VLM-based disaster damage severity classification. Task 1 retrieves the most likely georeferenced SVI--RSI candidates for each VGI query image through deterministic or probabilistic cross-view matching and multi-source score fusion. Task 2 uses VLMs to validate and interpret retrieved VGI--SVI--RSI candidate pairs through structured spatial reasoning based on visual appearance, spatial layout, and landmark evidence. It produces reasoning-supported matching decisions and interpretable spatial reasoning scores. Task 3 assesses disaster severity from VGI images based on six visually observable damage indicators, producing indicator-level scores, an overall severity level, and a brief rationale. The three tasks provide a unified framework for geolocalization, explainable match verification, and damage-aware disaster scene understanding.


\subsection{Deterministic and probabilistic cross-view geolocalization}

\begin{figure*}
    \centering
    \includegraphics[width=1\linewidth]{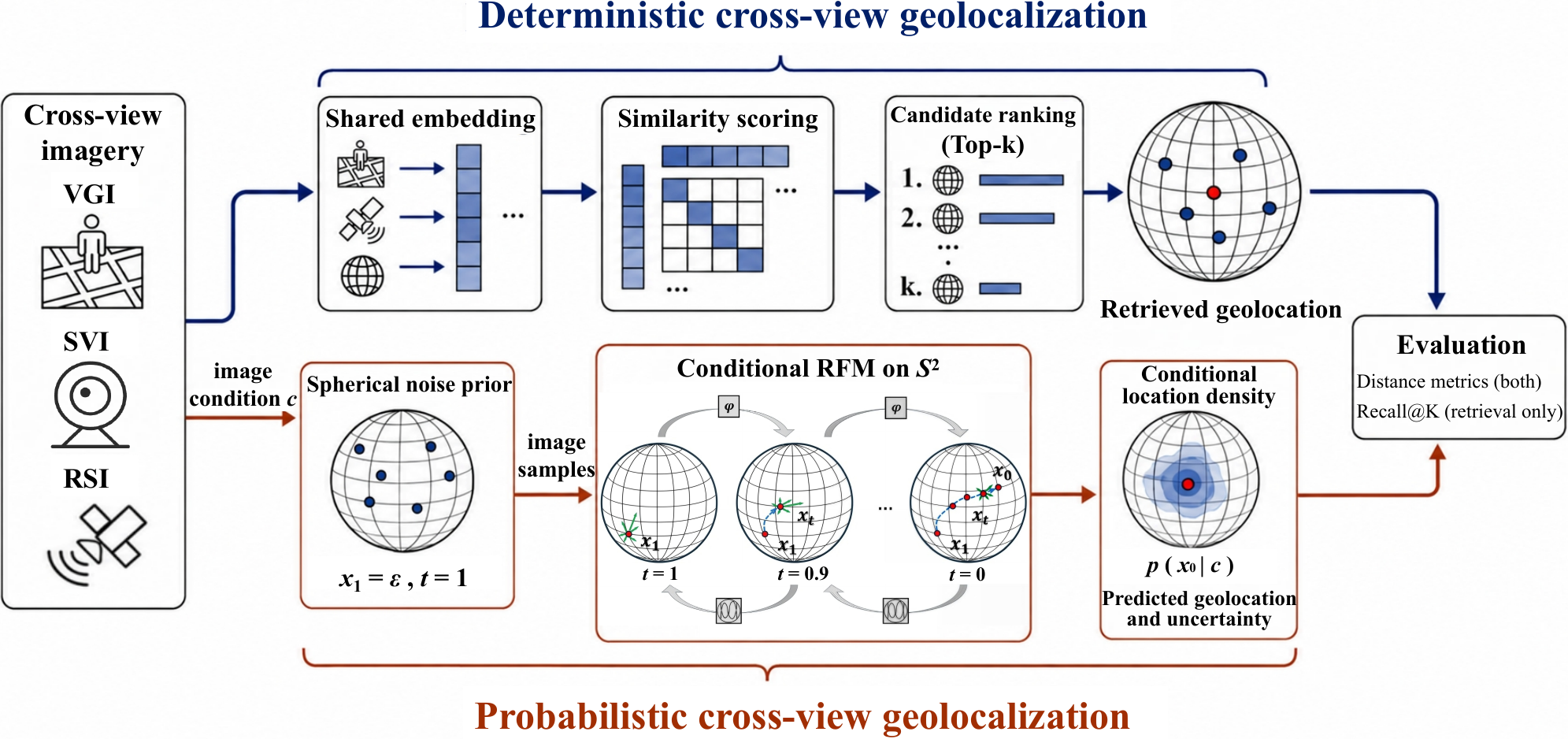}
    \caption{Overview of deterministic and probabilistic cross-view geolocalization.}
    \label{fig:CVGL}
\end{figure*}

We evaluate deterministic and probabilistic cross-view geolocalization methods on the disaster-oriented PhotoMappers benchmark. Each geolocation contains co-located multi-view observations, including ground-level disaster imagery (VGI), street-view panoramas (SVI), and RSI. Given a VGI query, the objective is to recover its geolocation by exploiting complementary visual evidence across these views. As shown in Fig.~\ref{fig:CVGL}, deterministic methods retrieve geolocations from a georeferenced reference gallery, whereas probabilistic methods estimate a conditional geographic distribution and its spatial uncertainty. The retrieval settings include VGI$\rightarrow$RSI, VGI$\rightarrow$SVI, SVI$\rightarrow$RSI, and multi-view configurations that jointly use VGI, SVI, and RSI.

For deterministic geolocalization, we evaluate SAIG-D \citep{zhu2023simple}, TransGeo \citep{zhu2022transgeo}, and Sample4Geo \citep{deuser2023sample4geo} with ConvNeXt-B \citep{liu2022convnet}, DINOv2 \citep{oquab2024dinov2}, and DINOv3 \citep{simeoni2025dinov3} visual backbones. The DINO-based models include base and large variants, denoted as DINOv2-B/L and DINOv3-B/L. Each image is encoded into a shared embedding space, and the similarity between a query embedding $L_c$ and a candidate reference embedding $L_r^i$ is measured using cosine similarity:

\begin{equation}
\label{eq:cosine_similarity}
S(L_c,L_r^i)=
\frac{L_c \cdot L_r^i}
{\|L_c\|\|L_r^i\|}
\end{equation}

Candidate references are ranked by their similarity scores, and the highest-ranked reference provides the predicted geolocation. Sinkhorn-based re-ranking is optionally applied to the DINOv3-B and DINOv3-L fusion models under the one-to-one matching setting. Independent ranking may produce hub references that receive high similarity scores from multiple queries. Sinkhorn--Knopp normalization \citep{sinkhorn1967concerning,cuturi2013sinkhorn} converts the query--reference similarity matrix into a doubly stochastic assignment matrix, suppressing hub references and promoting globally consistent matches without additional training.

For probabilistic geolocalization, we evaluate Riemannian flow-matching (RFM) conditioned on DINOv2-L and StreetCLIP representations, following the ProbGLC \citep{dufour2025around, li2026towards}. Unlike retrieval-based methods that select a geolocation from a finite reference gallery, RFM estimates a conditional distribution $p_{\varphi}(\mathbf{x}\mid\mathbf{c})$, where $\mathbf{x}\in\mathbb{S}^{2}$ denotes a geolocation on the spherical Earth surface and $\mathbf{c}$ is the visual condition. Both the ground-truth geolocation $\mathbf{x}_0$ and the noise $\epsilon$ lie on $\mathbb{S}^{2}$ \citep{chen2024flow}. The intermediate geolocation is defined along a spherical geodesic through the exponential and logarithmic maps:

\begin{equation}
\mathbf{x}_t =
\exp_{\mathbf{x}_{t-1}}\!\left(
\beta(t)\,
\log_{\mathbf{x}_{t-1}}(\epsilon)
\right)
\label{eq:rfm_spherical_path}
\end{equation}
where $\log_{\mathbf{x}_{t-1}}$ maps $\epsilon$ to the tangent space at $\mathbf{x}_{t-1}$, $\exp_{\mathbf{x}_{t-1}}$ maps the tangent vector back to $\mathbb{S}^{2}$, and $\beta(t)$ controls the interpolation. The conditional vector field $\varphi(\mathbf{x}_t\mid\mathbf{c})$ is trained to approximate the target velocity field $v(\mathbf{x}_t)$ by minimizing

\begin{equation}
\mathcal{L}_{\mathrm{RFM}}
=
\mathbb{E}_{\mathbf{x}_0,\mathbf{c},\epsilon,t}
\left[
\left\|
\varphi(\mathbf{x}_t\mid\mathbf{c})
-
v(\mathbf{x}_t)
\right\|_{\mathbf{x}_t}^{2}
\right]
\label{eq:rfm_loss}
\end{equation}
where $\lVert\cdot\rVert_{\mathbf{x}_t}$ is the norm induced by the Riemannian metric on the tangent space at $\mathbf{x}_t$.

An ordinary differential equation solver is initialized at a random geolocation $\mathbf{x}_1=\epsilon$ and integrates the learned vector field backward from $t=1$ to $t=0$. Exponential-map updates keep the trajectory on the spherical manifold, while repeated sampling produces multiple plausible geolocations and approximates the conditional geographic distribution. The spatial ambiguity of this distribution is summarized by the geolocalizability score \citep{dufour2025around, li2026towards}:

\begin{equation}
\mathrm{Localizability}(\mathbf{c})
=
\int_{\mathbb{S}^{2}}
p(\mathbf{x}_n\mid\mathbf{c})
\log_{2}P(\mathbf{x}_n\mid\mathbf{c})
\,d\mathbf{x}_n
\label{eq:localizability}
\end{equation}

where $\mathbf{x}_n$ denotes a sampled geolocation and the score corresponds to the negative entropy of the predicted distribution. A more concentrated distribution indicates stronger geographic evidence and lower spatial ambiguity, whereas a dispersed distribution reflects limited, ambiguous, or visually degraded cues in the disaster image.

Deterministic and probabilistic methods are compared using mean and median geolocalization errors and accuracy at distance thresholds of 1, 25, 200, 750, and 2,500~km. Retrieval-based methods are additionally evaluated using Recall@K at several ranking levels. These metrics provide a consistent basis for comparing methods that rank georeferenced candidates with those that directly estimate geographic coordinates.

\subsection{VLM-based cross-view geolocalization reasoning}
\subsubsection{Design of cross-view geolocalization reasoning indicators}

\begin{table*}[htbp]
\centering
\caption{Evaluation indicators for VLM-based interpretive reasoning of cross-view geolocalization results.}
\label{tab:vgi_svi_rsi_indicators_long}

\scriptsize
\renewcommand{\arraystretch}{1}
\setlength{\tabcolsep}{1.5pt}

\begin{tabular}{
>{\centering\arraybackslash}m{0.02\textwidth}
>{\raggedright\arraybackslash}m{0.42\textwidth}
>{\raggedright\arraybackslash}m{0.35\textwidth}
>{\centering\arraybackslash}m{0.06\textwidth}
>{\centering\arraybackslash}m{0.06\textwidth}
>{\centering\arraybackslash}m{0.06\textwidth}
}

\hline
\multirow[c]{2}{*}[-1.4ex]{\textbf{No.}} &
\multirow[c]{2}{*}[-1.4ex]{\textbf{Indicator name}} &
\multirow[c]{2}{*}[-1.4ex]{\textbf{Evaluation focus}} &
\multicolumn{3}{c}{\textbf{Dimension}} \\
\cline{4-6}
& & &
\scriptsize\textbf{Appearance} &
\scriptsize\textbf{Spatial layout} &
\scriptsize\textbf{Landmark cue} \\
\hline

1 & Building type and scale 
& Building and structure: Building appearance consistency 
& $\checkmark$ & & \\

2 & Overall building appearance 
& Building and structure: Building appearance consistency 
& $\checkmark$ & & \\

3 & Building position relative to nearby roads or open space 
& Building and structure: Building spatial-position consistency 
& & $\checkmark$ & \\

4 & Building position and shape correspondence in SVI--RSI 
& Building and structure: Building spatial-position consistency 
& & $\checkmark$ & \\

5 & Recognizable building outline 
& Building and structure: Corresponding building landmark 
& & & $\checkmark$ \\

6 & Distinctive building appearance or visible name 
& Building and structure: Corresponding building landmark 
& & & $\checkmark$ \\

\hline

7 & Road type and width 
& Road and transport: Road appearance consistency 
& $\checkmark$ & & \\

8 & Road markings and boundary features 
& Road and transport: Road appearance consistency 
& $\checkmark$ & & \\

9 & Road direction and connectivity 
& Road and transport: Road spatial-structure consistency 
& & $\checkmark$ & \\

10 & Road position relative to surrounding elements 
& Road and transport: Road spatial-structure consistency 
& & $\checkmark$ & \\

11 & Distinctive intersection, bridge, or road curve 
& Road and transport: Corresponding road landmark 
& & & $\checkmark$ \\

12 & Traffic lights, road signs, or road markers 
& Road and transport: Corresponding road landmark 
& & & $\checkmark$ \\

\hline

13 & Vegetation coverage and type 
& Vegetation and terrain: Natural-surface appearance consistency 
& $\checkmark$ & & \\

14 & Ground surface characteristics 
& Vegetation and terrain: Natural-surface appearance consistency 
& $\checkmark$ & & \\

15 & Vegetation position relative to roads or buildings 
& Vegetation and terrain: Natural-element spatial consistency 
& & $\checkmark$ & \\

16 & Waterbody, shoreline, or open-space position 
& Vegetation and terrain: Natural-element spatial consistency 
& & $\checkmark$ & \\

17 & Prominent trees, forest belts, or distinctive vegetation 
& Vegetation and terrain: Corresponding natural-terrain landmark 
& & & $\checkmark$ \\

18 & Coastline, river, lake, or slope outline 
& Vegetation and terrain: Corresponding natural-terrain landmark 
& & & $\checkmark$ \\

\hline

19 & Utility poles, streetlights, and cables 
& Objects and facilities: Stable facility appearance consistency 
& $\checkmark$ & & \\

20 & Road signs, fences, and guardrails 
& Objects and facilities: Stable facility appearance consistency 
& $\checkmark$ & & \\

21 & Facility position relative to roads or buildings 
& Objects and facilities: Facility spatial-distribution consistency 
& & $\checkmark$ & \\

22 & Local facility grouping and arrangement 
& Objects and facilities: Facility spatial-distribution consistency 
& & $\checkmark$ & \\

23 & Large facilities such as antennas, water towers, or pylons 
& Objects and facilities: Corresponding facility landmark 
& & & $\checkmark$ \\

24 & Recognizable signs such as storefront signs, billboards, or gas-station signs 
& Objects and facilities: Corresponding facility landmark 
& & & $\checkmark$ \\

\hline

25 & Scene functional type 
& Global scene: Overall scene-type consistency 
& $\checkmark$ & & \\

26 & Spatial density and openness 
& Global scene: Overall scene-type consistency 
& $\checkmark$ & & \\

27 & Overall building--road--vegetation relationship 
& Global scene: Overall scene-structure consistency 
& & $\checkmark$ & \\

28 & Consistency between the VGI visible scene and the SVI--RSI overall scene 
& Global scene: Overall scene-structure consistency 
& & $\checkmark$ & \\

29 & Large building complex, community, or commercial area 
& Global scene: Corresponding regional background cue 
& & & $\checkmark$ \\

30 & Coastline, river, bridge, or regional background 
& Global scene: Corresponding regional background cue 
& & & $\checkmark$ \\

\hline

\end{tabular}
\end{table*}

We design an indicator-based spatial reasoning framework to evaluate the quality and validity of VLM-generated explanations for VGI--SVI--RSI cross-view geolocalization results. The indicators were developed through iterative visual inspection of disaster images across different disaster types and scene categories. Domain experts discussed and organized recurring visual characteristics and spatial relationships into three dimensions: appearance, spatial layout, and landmark cue (Table~\ref{tab:vgi_svi_rsi_indicators_long}). Repeated reviews were conducted to identify missing cues and improve indicator coverage. These dimensions are further applied to five semantic classes, including building and structure, road and transport, vegetation and terrain, objects and facilities, and global scene. The resulting 30 indicators provide an initial, extensible set of stable visual and spatial cues. They reflect cues commonly used in human cross-view interpretation and support the assessment of spatial consistency between a VGI image and its candidate reference pair.

We further design a structured prompting scheme to guide VLMs in reasoning about and validating VGI--SVI--RSI cross-view geolocalization results. Following recent studies on structured prompting and visual reasoning evaluation \citep{wei2022chain,ye2025cross}, the prompt consists of four main components: task definition, output format, scoring scheme, and evaluation indicators. It defines the matching task between a post-disaster VGI image and its corresponding SVI--RSI reference pair, specifies the required structured output format, and introduces a three-level scoring scheme to quantify indicator-level reasoning quality. The 30 indicators are incorporated into the prompt to guide structured evaluation. To reduce temporal inconsistency, disaster-related damage information is excluded from the matching evidence, and the evaluation focuses on relatively stable geographic and visual characteristics shared across views. A containment-based matching criterion is used, where the VGI scene only needs to be reasonably supported by part of the SVI--RSI reference pair rather than requiring complete one-to-one correspondence. This design better reflects practical cross-view geolocalization scenarios with viewpoint differences and incomplete visual overlap.

\subsubsection{Reasoning and validating geolocalization results}

Spatial reasoning quality is evaluated using a criterion-based scoring scheme. For each spatial indicator, the generated reasoning is assigned a score of correct (2), partially correct (1), or incorrect (0), based on whether the indicator provides consistent visual evidence of a match between the VGI image and the corresponding SVI--RSI reference pair. A score of 2 indicates that the reasoning is directly supported by visual evidence from the VGI image and at least one reference view, without clear contradictions from other views. A score of 1 represents reasoning that is partially supported but affected by incomplete, weak, or ambiguous cross-view evidence. A score of 0 indicates that the relevant feature is not visible or matched across the compared views, or that the reasoning is unsupported, contradicted by visual evidence, spatially inconsistent, or hallucinated.



\begin{table*}[t]
\centering
\caption{Image-based indicators for VLM-based disaster damage severity reasoning and classification.}
\label{tab:damage_severity_indicators}

\scriptsize
\renewcommand{\arraystretch}{1.1}
\setlength{\tabcolsep}{4pt}

\begin{tabular}{
>{\raggedright\arraybackslash}m{0.19\textwidth}
>{\raggedright\arraybackslash}m{0.25\textwidth}
>{\raggedright\arraybackslash}m{0.25\textwidth}
>{\raggedright\arraybackslash}m{0.25\textwidth}
}

\hline
\multirow{2}{*}{\centering\textbf{Indicator name}} &
\multicolumn{3}{c}{\textbf{Damage severity}} \\
\cline{2-4}
& \textbf{Low} & \textbf{Medium} & \textbf{High} \\
\hline

Flooding and water impact 
& \textcolor{blue}{\textbf{Shallow}} or geolocalized water with \textcolor{blue}{\textbf{limited}} impact on visible objects. 
& Clear \textcolor{orange}{\textbf{flooding}} \textcolor{orange}{\textbf{affects}} roads, buildings, vehicles, or public spaces. 
& \textcolor{red}{\textbf{Deep}} water, strong flow, storm surge, or \textcolor{red}{\textbf{widespread}} inundation. \\

\hline

Building and structural damage 
& Buildings are mostly \textcolor{blue}{\textbf{intact}}, with no clear or only \textcolor{blue}{\textbf{minor}} damage. 
& \textcolor{orange}{\textbf{Partial}} damage is visible, such as doors, windows, walls, roofs, or water \textcolor{orange}{\textbf{intrusion}}. 
& Severe structural damage, \textcolor{red}{\textbf{collapse}}, heavy inundation, or \textcolor{red}{\textbf{loss}} of building function. \\

\hline

Road and access disruption 
& Roads or paths remain mostly \textcolor{blue}{\textbf{usable}}, with \textcolor{blue}{\textbf{minor}} debris or shallow water. 
& Roads are partially \textcolor{orange}{\textbf{blocked}} or \textcolor{orange}{\textbf{difficult}} to access due to water, debris, vehicles, or fallen trees. 
& Roads are \textcolor{red}{\textbf{impassable}} or heavily \textcolor{red}{\textbf{disrupted}} by deep water, large debris, collapse, or severe surface damage. \\

\hline

Vegetation and natural damage 
& Vegetation remains \textcolor{blue}{\textbf{intact}}, with only \textcolor{blue}{\textbf{small}} branches or geolocalized disturbance. 
& \textcolor{orange}{\textbf{Fallen}} trees, broken branches, or damaged vegetation \textcolor{orange}{\textbf{partially}} affect roads, buildings, or open spaces. 
& \textcolor{red}{\textbf{Severe}} vegetation damage, large fallen trees, or widespread natural disturbance \textcolor{red}{\textbf{blocks}} access or damages objects. \\

\hline

Infrastructure and facility damage 
& Facilities are mostly \textcolor{blue}{\textbf{intact}}, with only \textcolor{blue}{\textbf{minor}} or geolocalized disturbance. 
& Some facilities are \textcolor{orange}{\textbf{damaged}} or \textcolor{orange}{\textbf{disrupted}}, such as poles, signs, fences, utilities, or public equipment. 
& Critical infrastructure or multiple facilities are severely damaged, submerged, \textcolor{red}{\textbf{collapsed}}, or \textcolor{red}{\textbf{nonfunctional}}. \\

\hline

Debris and scene disruption 
& \textcolor{blue}{\textbf{Limited}} debris or scattered small objects, with the scene mostly \textcolor{blue}{\textbf{organized}}. 
& \textcolor{orange}{\textbf{Noticeable}} debris or displaced objects cause local \textcolor{orange}{\textbf{disorder}} across the scene. 
& \textcolor{red}{\textbf{Extensive}} debris, widespread destruction, or \textcolor{red}{\textbf{chaotic}} conditions affect scene elements. \\

\hline

\end{tabular}
\end{table*}

All VLM-based annotation and evaluation are conducted in a zero-shot setting without task-specific examples. Gemini-3.6-Flash \citep{doshi2026gemini36flash} is used as the scalable annotation and VLM-as-a-Judge model for its strong performance and cost efficiency, ranking second on the English-language Vision Arena Leaderboard \citep{arena2026vision}. The generated annotations serve as a scalable alternative to exhaustive manual labeling. For human-in-the-loop validation, 20\% of the samples are stratified according to predicted damage severity levels and independently reviewed by two human experts. The experts examine whether the predicted spatial matching decisions and corresponding reasoning are visually grounded in the VGI, SVI, and RSI evidence. Disagreements are resolved through discussion, providing additional quality control for identifying unsupported explanations, ambiguous cases, and systematic reasoning errors across VLMs. LLM/VLM-as-a-Judge frameworks with human-in-the-loop verification have been increasingly explored in multimodal understanding tasks \citep{wu2022survey,lin2026act}.

The evaluation of both spatial reasoning in cross-view geolocalization and disaster damage severity assessment follows the same evaluation framework. Qwen3.6-27B \citep{qwen2026qwen36}, GPT-4o \citep{openai2024gpt4o}, and InternVL3-8B \citep{zhu2025internvl3} are evaluated based solely on the visual evidence contained in each VGI image. Indicator-level scores and overall predictions are assessed using agreement rate and quadratic weighted kappa ($\kappa_w$). Compared with conventional Cohen's kappa, quadratic weighted kappa considers the ordinal relationships among scoring categories, assigning smaller penalties to near-miss disagreements and larger penalties to more substantial deviations \citep{cohen1968weighted}. The quadratic weighted kappa is defined as follows \citep{cohen1968weighted}:

\begin{equation}
\kappa_w = 1 - \frac{\sum_{i=1}^{K}\sum_{j=1}^{K} w_{ij} O_{ij}}
{\sum_{i=1}^{K}\sum_{j=1}^{K} w_{ij} E_{ij}}
\end{equation}

where \(K\) is the number of scoring categories, \(O_{ij}\) denotes the observed agreement matrix between VLM and human scores, and \(E_{ij}\) represents the expected agreement matrix estimated from the marginal distributions. The quadratic weighting matrix $w_{ij}$ is defined as:

\begin{equation}
w_{ij}=\frac{(i-j)^2}{(K-1)^2}
\end{equation}

where \(i\) and \(j\) represent the assigned score categories. This weighting scheme penalizes larger disagreements more severely while assigning lower penalties to adjacent categories.

The observed agreement matrix is calculated as:

\begin{equation}
O_{ij}=\sum_{n=1}^{N}\mathbf{1}(\hat{s}_n=i)\mathbf{1}(s_n=j),
\end{equation}

where \(N\) is the number of evaluated samples, \(\hat{s}_n\) is the VLM-generated score for the \(n\)-th sample, \(s_n\) is the corresponding human-verified score, and \(\mathbf{1}(\cdot)\) is an indicator function. The expected agreement matrix is obtained from the marginal distributions:

\begin{equation}
E_{ij}=\frac{n_i^{\mathrm{VLM}} \cdot n_j^{\mathrm{Human}}}{N}
\end{equation}

where \(n_i^{\mathrm{VLM}}\) and \(n_j^{\mathrm{Human}}\) denote the numbers of samples assigned to categories \(i\) and \(j\) by the VLM and human annotators, respectively.

The scoring categories are defined as \(C=\{0,1,2\}\), corresponding to the three-level damage severity or reasoning quality scores. The quadratic weighted kappa therefore measures the agreement between VLM predictions and human annotations while accounting for both chance agreement and the ordinal distance between different scoring categories.

\subsection{VLM-based disaster damage reasoning and classification}

We classify the damage severity of collected VGI images into three levels: low, medium, and high, using six visually observable indicators: flooding and water impact, building and structural damage, road and access disruption, vegetation and natural damage, infrastructure and facility damage, and debris and scene disruption. As shown in Table~\ref{tab:damage_severity_indicators}, these indicators capture different aspects of disaster impacts and enable both indicator-level assessment and overall severity classification.

All VLM-based annotation generation and evaluation are conducted in a zero-shot setting without task-specific examples. We evaluate Qwen3.6-27B \citep{qwen2026qwen36}, GPT-4o \citep{openai2024gpt4o}, and InternVL3-8B \citep{zhu2025internvl3} using only the visual information in each VGI image. For each image, the models generate scores for individual damage indicators, an overall severity level, and reasoning explanations. As described above, Gemini-3.6-Flash \citep{doshi2026gemini36flash} serves as the scalable VLM-as-a-Judge for annotation, while 20\% of samples are stratified by predicted severity and independently verified by two human experts in a human-in-the-loop process. The experts assess the same indicators and determine whether the predicted scores, severity levels, and reasoning are supported by visible image evidence. Following the human-aligned verification strategy used for spatial reasoning evaluation, two experts independently assess VLM outputs and resolve disagreements through discussion to obtain final verified labels \citep{lin2026act,hou2027postgiser}. Agreement rate and quadratic weighted kappa ($\kappa_w$) \citep{cohen1968weighted} measure consistency between VLM predictions and human verification while accounting for the ordinal relationships among severity categories.

\section{Experiments} \label{Experiment}

\subsection{Dataset overview and experimental settings}

\begin{figure*}
    \centering
    \includegraphics[width=\linewidth]{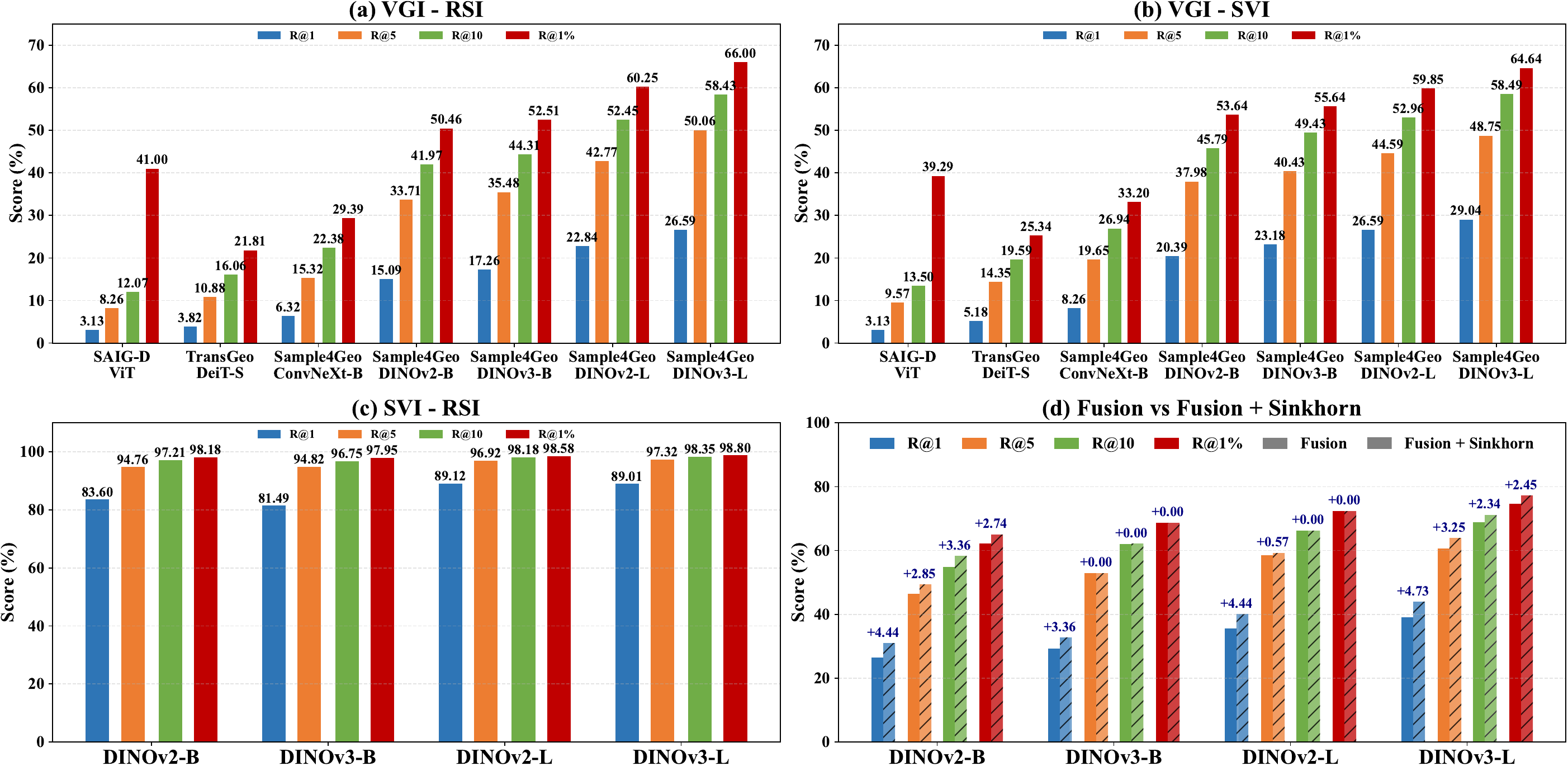}
    \caption{Deterministic cross-view geolocalization performance under different retrieval and fusion settings: (a) VGI--RSI retrieval, (b) VGI--SVI retrieval, (c) SVI--RSI retrieval, and (d) VGI--SVI--RSI fusion with and without Sinkhorn re-ranking.}
    \label{fig:CVGL_results}
\end{figure*}

The experiments are conducted on a curated cross-view PhotoMappers dataset. The original collection contained more than 27,575 crowdsourced disaster images, including duplicate copies of the same image with identical latitude and longitude coordinates but different image dimensions. After retaining one record from each duplicate group, approximately 13,513 candidate VGI records remained. Georeferenced SVI and RSI were subsequently collected for these locations, producing more than 9,000 candidate cross-view groups. Irrelevant, uninformative, and visually unusable samples were further removed through manual inspection. The final dataset contains 26,340 images organized into 8,780 validated VGI--SVI--RSI triplets for evaluating cross-view geolocalization, spatial reasoning, and disaster damage assessment, as shown in  Fig. \ref{fig:dataset}.

For Task 1, cross-view geolocalization, we adopt a standard 8:2 train--test split. The training set contains 7,024 geolocation groups, corresponding to 21,072 images in the three-view VGI--SVI--RSI setting or 14,048 images in each two-view setting. The test set contains 1,756 groups, corresponding to 5,268 three-view images or 3,512 two-view images. Its reference gallery comprises 1,756 distinct geolocations for each reference-view configuration. Samples are sorted by latitude and longitude and then uniformly assigned to the two subsets, ensuring that no geolocation is shared between the training and test sets. The VLM-based spatial reasoning and validation tasks are evaluated zero-shot, without task-specific examples or model fine-tuning.

\subsection{Cross-view geolocalization performance}

\begin{table*}
\centering
\caption{Comparison of geolocalization performance between generative and retrieval-based approaches. Median and mean are errors in kilometers, while @1,km--@2500,km are the percentages of predictions within each distance threshold.}
\label{tab:generative_retrieval_comparison}
\small
\renewcommand{\arraystretch}{1.10}
\setlength{\tabcolsep}{5pt}
\begin{tabular}{
>{\centering\arraybackslash}m{3cm}   
>{\centering\arraybackslash}m{1cm}   
>{\centering\arraybackslash}m{3.5cm} 
>{\centering\arraybackslash}m{1.2cm} 
>{\centering\arraybackslash}m{1.0cm} 
>{\centering\arraybackslash}m{1.0cm} 
>{\centering\arraybackslash}m{1.0cm} 
>{\centering\arraybackslash}m{1.0cm} 
>{\centering\arraybackslash}m{1.0cm} 
>{\centering\arraybackslash}m{1.0cm} 
}
\toprule

\textbf{Method}
& \textbf{Type}
& \textbf{View Input}
& \textbf{Median (km)}
& \textbf{Mean (km)}
& \textbf{@1km}
& \textbf{@25km}
& \textbf{@200km}
& \textbf{@750km}
& \textbf{@2500km} \\
\midrule

DINOv3-B
& \textcolor{blue}{$\boldsymbol{\times}$}
& VGI--RSI
& 381.47
& 747.35
& 19.13
& 23.29
& 37.81
& 68.17
& 93.96 \\

DINOv3-L
& \textcolor{blue}{$\boldsymbol{\times}$}
& VGI--RSI
& 218.67
& 525.35
& 28.25
& 33.20
& 48.35
& 76.25
& 96.18 \\

DINOv3-B
& \textcolor{blue}{$\boldsymbol{\times}$}
& VGI--SVI
& 340.37
& 631.42
& 25.11
& 28.59
& 40.43
& 70.39
& 95.10 \\

DINOv3-L
& \textcolor{blue}{$\boldsymbol{\times}$}
& VGI--SVI
& 193.99
& 545.46
& 31.44
& 35.71
& 50.34
& 75.40
& 96.58 \\

DINOv3-B
& \textcolor{blue}{$\boldsymbol{\times}$}
& VGI--SVI--RSI
& 206.96
& 547.83
& 31.78
& 35.99
& 49.43
& 74.94
& 95.90 \\

DINOv3-L
& \textcolor{blue}{$\boldsymbol{\times}$}
& VGI--SVI--RSI
& \textbf{65.31}
& \textbf{390.08}
& \textbf{41.40}
& \textbf{45.62}
& \textbf{58.60}
& \textbf{81.44}
& \textbf{98.35} \\

DINOv3-B
& \textcolor{blue}{$\boldsymbol{\times}$}
& VGI--SVI--RSI + Sinkhorn
& 204.06
& 567.15
& 34.40
& 37.98
& 49.66
& 74.94
& 95.73 \\

DINOv3-L
& \textcolor{blue}{$\boldsymbol{\times}$}
& VGI--SVI--RSI + Sinkhorn
& \textbf{27.29}
& \textbf{359.49}
& \textbf{45.96}
& \textbf{49.72}
& \textbf{62.19}
& \textbf{82.12}
& \textbf{98.12} \\

RFM DINOv2-L
& \textcolor{red}{$\boldsymbol{\checkmark}$}
& None
& 324.17
& 651.50
& 0.00
& 9.45
& 39.75
& 71.58
& 95.50 \\

RFM StreetCLIP
& \textcolor{red}{$\boldsymbol{\checkmark}$}
& None
& \textbf{129.34}
& \textbf{370.66}
& \textbf{0.06}
& \textbf{10.76}
& \textbf{60.93}
& \textbf{86.56}
& \textbf{98.23} \\

\bottomrule
\end{tabular}

\vspace{1ex}
\raggedright
\footnotesize \textit{Note:} \textcolor{blue}{$\boldsymbol{\times}$} indicates retrieval-based approaches, while \textcolor{red}{$\boldsymbol{\checkmark}$} indicates generative approaches.
\end{table*}

\begin{figure*}
    \centering
    \includegraphics[width=\linewidth]{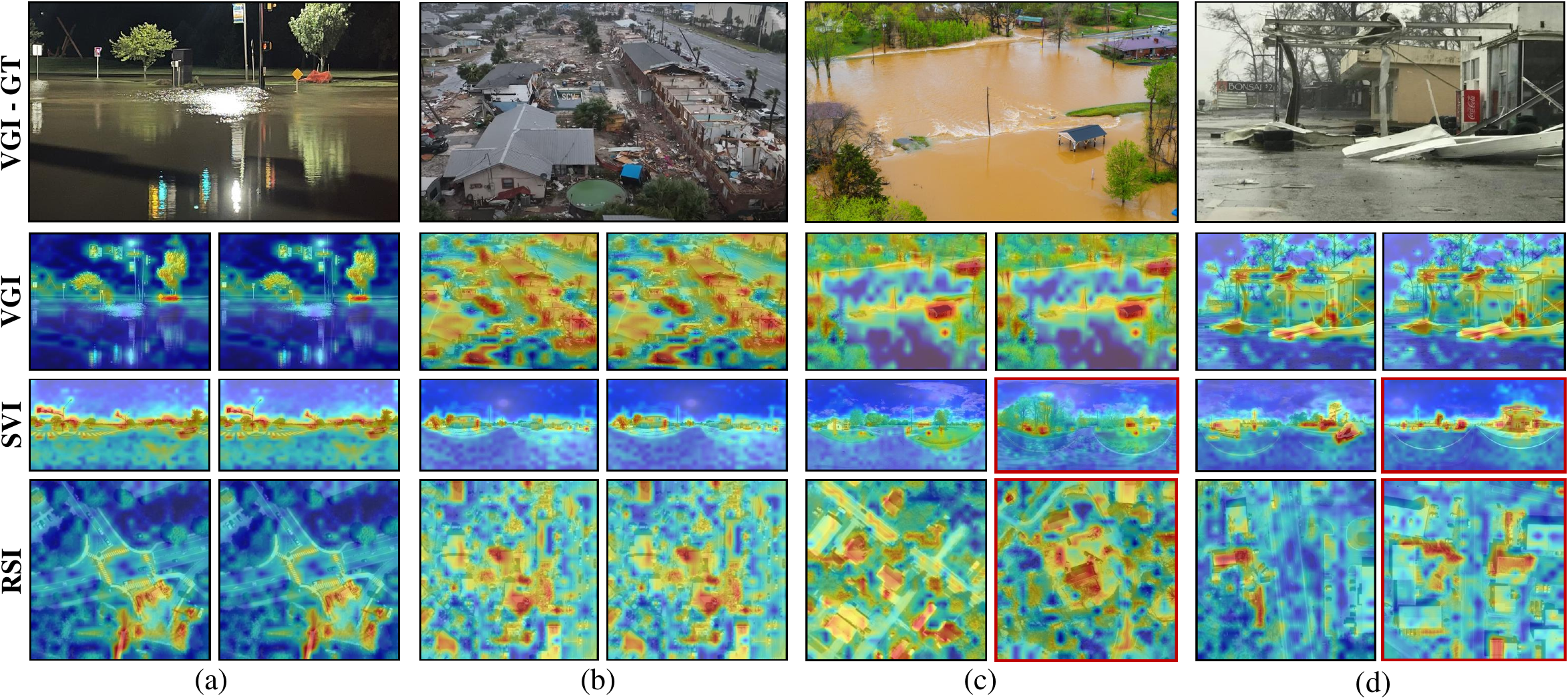}
    \vspace{-15pt}
    \caption{Heatmaps for (a) and (b) correctly and (c) and (d) incorrectly geolocalized imagery pairs generated by our best-performing DINOv3-L model with VGI--SVI--RSI fusion and Sinkhorn re-ranking, with the corresponding raw VGI ground-truth images included for reference. The first column shows the ground-truth images, and the second column shows the predicted results.}
    \label{fig:/CVGL_heatmap}
\end{figure*}

We evaluate cross-view geolocalization performance on the proposed PhotoMappers dataset under different retrieval settings, including VGI-to-RSI, VGI-to-SVI, and multi-view fusion. Recall at different ranks is used for evaluation, where R@K denotes the proportion of queries whose correct reference appears among the top-$K$ retrieved candidates. As shown in Fig. \ref{fig:CVGL_results}, existing methods exhibit relatively limited geolocalization capability when directly handling the large visual domain gap between ground-level and aerial/satellite views. Conventional retrieval-based approaches, including SAIG-D \citep{zhu2023simple}, TransGeo \citep{zhu2022transgeo}, and Sample4Geo \citep{deuser2023sample4geo} with ConvNeXt-B \citep{liu2022convnet}, achieve relatively low R@1 performance in the VGI$\rightarrow$RSI setting, obtaining only 3.13\%, 3.82\%, and 6.32\%, respectively. This indicates that their visual representations struggle to capture discriminative cross-view semantic correspondences under substantial viewpoint and appearance variations.

Replacing the backbone with stronger foundation models brings consistent improvements across all evaluation settings. With DINOv2 \citep{oquab2024dinov2} and DINOv3 \citep{simeoni2025dinov3} features, Sample4Geo achieves significant gains compared with previous baselines. In the VGI$\rightarrow$RSI task, DINOv3-L improves R@1 from 6.32\% to 26.59\% and increases R@1\% to 66.00\%. Similar trends can be observed in the VGI$\rightarrow$SVI scenario, where DINOv3-L achieves 29.04\% R@1 and 64.64\% R@1\%. The gains across different retrieval levels indicate that stronger visual representations capture more reliable semantic and structural features for matching disaster images across viewpoints.

The SVI$\rightarrow$RSI evaluation further highlights the effectiveness of the cross-view representations. Compared with the VGI-based retrieval tasks, SVI and RSI preserve more consistent scene structures and geographic patterns, resulting in substantially higher retrieval accuracy. All DINO-based variants perform strongly, with DINOv3-L reaching 89.01\% R@1, 97.32\% R@5, 98.35\% R@10, and 98.80\% R@1\%. These results show that roads, building configurations, and surrounding spatial layouts can be matched more reliably when the cross-view discrepancy is reduced. The substantial performance gap between VGI-based retrieval and SVI$\rightarrow$RSI retrieval further identifies disaster VGI as a key bottleneck in the geolocalization pipeline. Even with DINOv3-L, VGI$\rightarrow$RSI and VGI$\rightarrow$SVI achieve only 26.59\% and 29.04\% R@1, respectively, compared with 89.01\% for SVI$\rightarrow$RSI. This gap can be attributed to disaster-induced appearance changes, unconstrained viewpoints, occlusion, limited fields of view, and temporal differences between VGI and reference imagery. For practical deployment, improving the initial linkage between disaster VGI and georeferenced references is therefore critical, motivating multi-view fusion, uncertainty-aware prediction, and additional verification for low-confidence matches.

The VGI--SVI--RSI multi-view fusion strategy further improves geolocalization accuracy by integrating complementary information from different observation perspectives. As shown in Table \ref{tab:generative_retrieval_comparison}, the fusion model consistently achieves higher retrieval performance than individual-view matching. For DINOv3-L, the fusion of VGI, SVI, and RSI improves R@1 over the single-view results to 39.12\%, while adding the Sinkhorn-based alignment module further increases R@1 to 43.85\%. Similar improvements are observed across other retrieval thresholds, with R@1\% increasing from 74.66\% to 77.11\%. These results suggest that multi-view feature interaction provides more robust and complementary geographic cues while reducing ambiguity caused by individual image sources.  The heatmaps in Fig. \ref{fig:/CVGL_heatmap} show that correct matches rely on consistent structural cues across views, whereas incorrect matches are affected by ambiguous scene features.

We further compare retrieval-based methods with generative geolocalization methods on the newly curated PhotoMappers dataset. As shown in Table \ref{tab:generative_retrieval_comparison}, retrieval-based models achieve more stable performance across different distance thresholds. The deterministic DINOv3-L fusion model with Sinkhorn alignment obtains a median geolocalization error of only 27.29 km and reaches 98.12\% accuracy within 2500 km. Compared with the generative RFM DINOv2-L and RFM StreetCLIP baselines, the retrieval-based DINOv3-L model provides substantially better fine-grained geolocalization, particularly at strict distance thresholds. It achieves 45.96\% accuracy within 1~km, whereas the two RFM-based approaches obtain only 0.00\% and 0.06\%, respectively. These results indicate that deterministic cross-view retrieval with multi-view fusion and feature alignment is more effective for precise geolocalization on our newly curated disaster dataset.

\subsection{Cross-view spatial reasoning and validation results}

\begin{table*}[htbp]
\centering
\caption{Agreement and quadratic weighted kappa ($\kappa_w$) evaluation for cross-view geolocalization reasoning. Bold the best-performing results in each category. Bold the best-performing results in each category.}
\label{tab:vlm_indicator_agreement}
\footnotesize
\renewcommand{\arraystretch}{1.10}
\setlength{\tabcolsep}{3pt}

\resizebox{\textwidth}{!}{%
\begin{tabular}{
>{\centering\arraybackslash}p{0.33cm}
>{\raggedright\arraybackslash}p{4.5cm}
!{\vrule width 0.6pt}
cc cc cc
!{\vrule width 0.6pt}
cc cc cc
}
\toprule

\multirow{2}{*}{\textbf{No.}}
& \multirow{2}{*}{\textbf{Indicator Abbr.}}
& \multicolumn{2}{c}{\textbf{VLM vs Exp. A}}
& \multicolumn{2}{c}{\textbf{VLM vs Exp. B}}
& \multicolumn{2}{c!{\vrule width 0.6pt}}{\textbf{Exp. A vs Exp. B}}
& \multicolumn{2}{c}{\textbf{Qwen}}
& \multicolumn{2}{c}{\textbf{GPT}}
& \multicolumn{2}{c}{\textbf{InternVL}} \\

\cmidrule(lr){3-4}
\cmidrule(lr){5-6}
\cmidrule(lr){7-8}
\cmidrule(lr){9-10}
\cmidrule(lr){11-12}
\cmidrule(lr){13-14}

&
& \textbf{Agr.} & \textbf{$\kappa_w$}
& \textbf{Agr.} & \textbf{$\kappa_w$}
& \textbf{Agr.} & \textbf{$\kappa_w$}
& \textbf{Agr.} & \textbf{$\kappa_w$}
& \textbf{Agr.} & \textbf{$\kappa_w$}
& \textbf{Agr.} & \textbf{$\kappa_w$} \\
\midrule

1 & Building type and scale
& 0.954 & 0.964 & 0.991 & 0.985 & 0.957 & 0.975
& 0.644 & 0.434 & \textbf{0.708} & \textbf{0.610} & 0.537 & 0.376 \\

2 & Overall building appearance
& 0.983 & 0.990 & 0.874 & 0.921 & 0.862 & 0.914
& 0.632 & 0.429 & \textbf{0.702} & \textbf{0.613} & 0.504 & 0.364 \\

3 & Building-road-open space position
& 0.940 & 0.965 & 0.991 & 0.995 & 0.948 & 0.970
& 0.637 & 0.407 & \textbf{0.694} & \textbf{0.585} & 0.517 & 0.355 \\

4 & Building shape correspondence
& 0.989 & 0.989 & 0.968 & 0.977 & 0.974 & 0.985
& 0.649 & 0.415 & \textbf{0.710} & \textbf{0.578} & 0.522 & 0.351 \\

5 & Recognizable building outline
& 0.991 & 0.990 & 0.991 & 0.990 & 0.994 & 0.997
& 0.633 & 0.426 & \textbf{0.724} & \textbf{0.630} & 0.556 & 0.396 \\

6 & Distinctive building appearance
& 0.966 & 0.930 & 0.943 & 0.920 & 0.974 & 0.984
& 0.587 & 0.426 & \textbf{0.672} & \textbf{0.558} & 0.529 & 0.375 \\

\midrule

7 & Road type and width
& 0.885 & 0.917 & 0.971 & 0.978 & 0.862 & 0.898
& 0.604 & 0.445 & \textbf{0.661} & \textbf{0.591} & 0.438 & 0.268 \\

8 & Road markings and boundaries
& 0.928 & 0.923 & 0.971 & 0.953 & 0.957 & 0.969
& 0.552 & 0.459 & \textbf{0.578} & \textbf{0.531} & 0.453 & 0.292 \\

9 & Road direction and connectivity
& 0.923 & 0.952 & 0.960 & 0.974 & 0.883 & 0.926
& 0.596 & 0.416 & \textbf{0.634} & \textbf{0.534} & 0.472 & 0.308 \\

10 & Road-element position relationship
& 0.960 & 0.973 & 0.997 & 0.998 & 0.963 & 0.975
& 0.669 & 0.443 & \textbf{0.708} & \textbf{0.555} & 0.462 & 0.267 \\

11 & Distinctive intersections or curves
& 0.989 & 0.979 & 0.994 & 0.983 & 0.994 & 0.996
& 0.530 & 0.393 & \textbf{0.620} & \textbf{0.545} & 0.473 & 0.256 \\

12 & Traffic signs and markers
& 0.980 & 0.954 & 0.983 & 0.956 & 0.991 & 0.994
& 0.604 & 0.481 & \textbf{0.613} & \textbf{0.531} & 0.498 & 0.348 \\

\midrule

13 & Vegetation coverage and type
& 0.822 & 0.821 & 0.994 & 0.994 & 0.817 & 0.814
& 0.626 & 0.379 & \textbf{0.686} & \textbf{0.502} & 0.407 & 0.173 \\

14 & Ground surface characteristics
& 0.837 & 0.845 & 0.968 & 0.971 & 0.822 & 0.843
& 0.571 & 0.407 & \textbf{0.604} & \textbf{0.504} & 0.418 & 0.223 \\

15 & Vegetation-road-building position
& 0.968 & 0.977 & 0.997 & 0.998 & 0.971 & 0.979
& 0.652 & 0.432 & \textbf{0.677} & \textbf{0.493} & 0.437 & 0.211 \\

16 & Waterbody and open-space position
& 0.943 & 0.960 & 0.874 & 0.903 & 0.817 & 0.866
& 0.513 & 0.324 & \textbf{0.518} & \textbf{0.345} & 0.399 & 0.126 \\

17 & Prominent vegetation features
& 0.946 & 0.948 & 0.954 & 0.953 & 0.940 & 0.956
& 0.563 & 0.408 & \textbf{0.576} & \textbf{0.451} & 0.414 & 0.212 \\

18 & Coastline and terrain outline
& 0.905 & 0.922 & 0.900 & 0.830 & 0.954 & 0.964
& 0.534 & 0.240 & \textbf{0.573} & \textbf{0.277} & 0.450 & 0.142 \\

\midrule

19 & Utility poles and cables
& 0.934 & 0.952 & 0.997 & 0.992 & 0.937 & 0.959
& 0.571 & 0.472 & \textbf{0.616} & \textbf{0.595} & 0.448 & 0.277 \\

20 & Road signs, fences, and guardrails
& 0.989 & 0.993 & 0.997 & 0.998 & 0.991 & 0.995
& 0.580 & 0.465 & \textbf{0.585} & \textbf{0.524} & 0.474 & 0.311 \\

21 & Facility-road-building position
& 0.974 & 0.985 & 0.968 & 0.981 & 0.943 & 0.965
& \textbf{0.637} & 0.410 & 0.614 & \textbf{0.414} & 0.474 & 0.287 \\

22 & Facility grouping arrangement
& 0.971 & 0.979 & 0.994 & 0.992 & 0.977 & 0.987
& \textbf{0.683} & \textbf{0.438} & 0.635 & 0.401 & 0.486 & 0.296 \\

23 & Large facility structures
& 0.871 & 0.855 & 0.862 & 0.842 & 0.991 & 0.990
& 0.589 & 0.112 & \textbf{0.643} & \textbf{0.181} & 0.510 & 0.111 \\

24 & Recognizable commercial signs
& 0.934 & 0.837 & 0.931 & 0.834 & 0.991 & 0.995
& 0.621 & 0.380 & \textbf{0.695} & \textbf{0.514} & 0.531 & 0.367 \\

\midrule

25 & Scene functional type
& 0.903 & 0.905 & 0.931 & 0.925 & 0.834 & 0.812
& 0.708 & 0.393 & \textbf{0.751} & \textbf{0.464} & 0.399 & 0.159 \\

26 & Spatial density and openness
& 0.885 & 0.889 & 0.997 & 0.997 & 0.883 & 0.885
& 0.682 & 0.385 & \textbf{0.731} & \textbf{0.454} & 0.392 & 0.166 \\

27 & Building-road-vegetation relationship
& 0.980 & 0.980 & 0.989 & 0.986 & 0.991 & 0.994
& \textbf{0.700} & 0.450 & 0.687 & \textbf{0.481} & 0.449 & 0.228 \\

28 & VGI-SVI-RSI scene consistency
& 0.989 & 0.993 & 0.983 & 0.989 & 0.971 & 0.982
& \textbf{0.721} & 0.419 & 0.681 & \textbf{0.450} & 0.407 & 0.186 \\

29 & Large building complexes
& 0.848 & 0.903 & 0.914 & 0.823 & 0.860 & 0.911
& 0.517 & 0.337 & \textbf{0.593} & \textbf{0.487} & 0.498 & 0.334 \\

30 & Coastline and regional background
& 0.946 & 0.936 & 0.977 & 0.955 & 0.963 & 0.976
& 0.478 & 0.241 & \textbf{0.489} & \textbf{0.262} & 0.409 & 0.128 \\



\specialrule{0.6pt}{2pt}{2pt}

\multicolumn{2}{l!{\vrule width 0.6pt}}{\textbf{Building and structure average}}
& 0.971 & 0.971 & 0.960 & 0.965 & 0.952 & 0.971
& 0.630 & 0.423 & \textbf{0.702} & \textbf{0.596} & 0.528 & 0.370 \\

\multicolumn{2}{l!{\vrule width 0.6pt}}{\textbf{Road and transport average}}
& 0.944 & 0.950 & 0.979 & 0.974 & 0.942 & 0.960
& 0.593 & 0.440 & \textbf{0.636} & \textbf{0.548} & 0.466 & 0.290 \\

\multicolumn{2}{l!{\vrule width 0.6pt}}{\textbf{Vegetation and terrain average}}
& 0.904 & 0.912 & 0.948 & 0.942 & 0.887 & 0.904
& 0.577 & 0.365 & \textbf{0.606} & \textbf{0.429} & 0.421 & 0.181 \\

\multicolumn{2}{l!{\vrule width 0.6pt}}{\textbf{Objects and facilities average}}
& 0.946 & 0.934 & 0.958 & 0.940 & 0.972 & 0.982
& 0.614 & 0.380 & \textbf{0.631} & \textbf{0.438} & 0.487 & 0.275 \\

\multicolumn{2}{l!{\vrule width 0.6pt}}{\textbf{Global scene average}}
& 0.925 & 0.934 & 0.965 & 0.946 & 0.917 & 0.927
& 0.634 & 0.371 & \textbf{0.655} & \textbf{0.433} & 0.426 & 0.200 \\

\midrule

\multicolumn{2}{l!{\vrule width 0.6pt}}{\textbf{Appearance dimension average}}
& 0.912 & 0.920 & 0.969 & 0.971 & 0.892 & 0.906
& 0.617 & 0.427 & \textbf{0.662} & \textbf{0.539} & 0.447 & 0.261 \\

\multicolumn{2}{l!{\vrule width 0.6pt}}{\textbf{Spatial layout dimension average}}
& 0.964 & 0.975 & 0.972 & 0.979 & 0.944 & 0.963
& 0.646 & 0.415 & \textbf{0.656} & \textbf{0.484} & 0.463 & 0.262 \\

\multicolumn{2}{l!{\vrule width 0.6pt}}{\textbf{Landmark cue dimension average}}
& 0.938 & 0.925 & 0.945 & 0.909 & 0.965 & 0.976
& 0.566 & 0.344 & \textbf{0.620} & \textbf{0.444} & 0.487 & 0.267 \\

\specialrule{0.9pt}{2pt}{2pt}

\multicolumn{2}{l!{\vrule width 0.6pt}}{\textbf{Overall indicator-level average}}
& 0.938 & 0.940
&  0.962 & 0.953
&  0.934 & 0.949
&  0.609 & 0.396
&  \textbf{0.646} & \textbf{0.489}
&  0.465 & 0.263 \\

\bottomrule
\end{tabular}
}
\end{table*}

\begin{figure*}
    \centering
    \includegraphics[width=\linewidth]{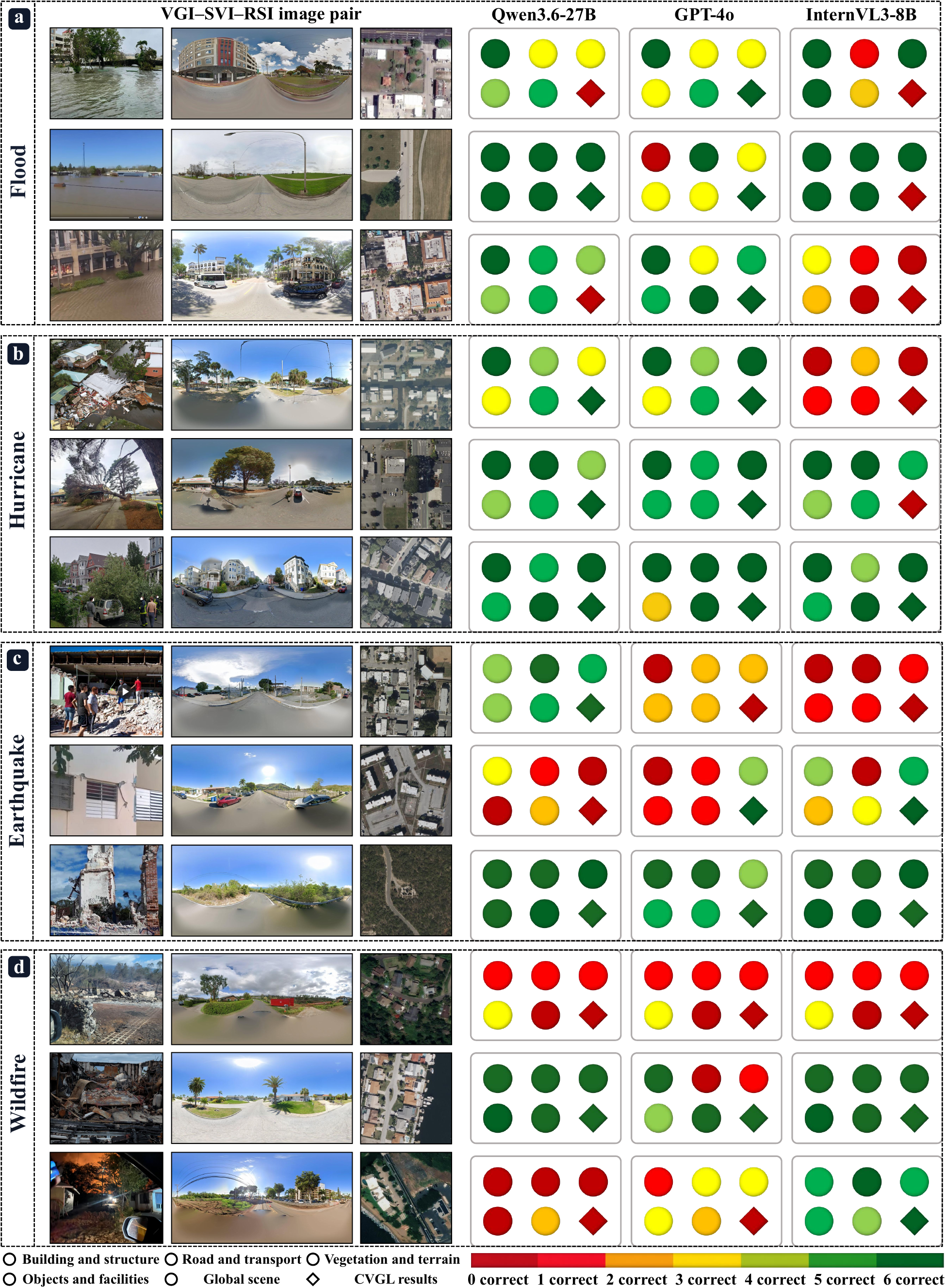}
    \vspace{-15pt}
    \caption{Spatial reasoning for cross-view geolocalization across disaster types. Three VLMs are compared across building and structure, road and transport, vegetation and terrain, objects and facilities, global scene (circles), and cross-view geolocalization (diamond), with different colors indicating reasoning results.}
    \label{fig:resultsCVGL}
\end{figure*}

\begin{table*}[t]
\centering
\caption{Agreement and quadratic weighted kappa ($\kappa_w$) evaluation between Gemini and two expert annotations for disaster damage severity assessment with 20\% expert verification.}
\label{tab:expert_vlm_damage_comparison}
\footnotesize
\renewcommand{\arraystretch}{1}
\setlength{\tabcolsep}{10pt}

\resizebox{\textwidth}{!}{%
\begin{tabular}{
>{\arraybackslash}m{3.5cm}
>{\arraybackslash}m{1cm}
!{\vrule width 0.6pt}
>{\centering\arraybackslash}p{1.2cm}
>{\centering\arraybackslash}p{1.2cm}
!{\vrule width 0.6pt}
>{\centering\arraybackslash}p{1.2cm}
>{\centering\arraybackslash}p{1.2cm}
!{\vrule width 0.6pt}
>{\centering\arraybackslash}p{1.2cm}
>{\centering\arraybackslash}p{1.2cm}
}
\toprule

\multirow{2}{*}{\textbf{Indicator}}
& \multirow{2}{*}{\textbf{Level}}
& \multicolumn{2}{c!{\vrule width 0.6pt}}{\textbf{VLM vs Expert A}}
& \multicolumn{2}{c!{\vrule width 0.6pt}}{\textbf{VLM vs Expert B}}
& \multicolumn{2}{c}{\textbf{Expert A vs Expert B}} \\
\cmidrule(lr){3-4} \cmidrule(lr){5-6} \cmidrule(lr){7-8}

&
& \multicolumn{1}{c}{\textbf{Agr.}} & \multicolumn{1}{c!{\vrule width 0.6pt}}{\textbf{$\kappa_w$}}
& \multicolumn{1}{c}{\textbf{Agr.}} & \multicolumn{1}{c!{\vrule width 0.6pt}}{\textbf{$\kappa_w$}}
& \multicolumn{1}{c}{\textbf{Agr.}} & \multicolumn{1}{c}{\textbf{$\kappa_w$}} \\
\midrule

\multirow{4}{*}{Flooding and water impact}
& Low
& 0.966 & 0.954 & 0.932 & 0.906 & 0.932 & 0.908 \\
& Medium
& 0.940 & 0.963 & 0.940 & 0.948 & 0.949 & 0.953 \\
& High
& 0.940 & 0.964 & 0.932 & 0.957 & 0.974 & 0.984 \\
& \textit{Overall}
& 0.949 & 0.968 & 0.934 & 0.953 & 0.952 & 0.964 \\

\midrule

\multirow{4}{*}{Building and structural damage}
& Low
& 0.983 & 0.916 & 0.983 & 0.927 & 0.991 & 0.846 \\
& Medium
& 0.949 & 0.948 & 0.957 & 0.931 & 0.915 & 0.912 \\
& High
& 0.932 & 0.948 & 0.966 & 0.974 & 0.915 & 0.936 \\
& \textit{Overall}
& 0.954 & 0.962 & 0.969 & 0.967 & 0.940 & 0.943 \\

\midrule

\multirow{4}{*}{Road and access disruption}
& Low
& 0.991 & 0.988 & 0.991 & 0.988 & 0.983 & 0.976 \\
& Medium
& 0.932 & 0.938 & 0.949 & 0.929 & 0.906 & 0.914 \\
& High
& 0.949 & 0.921 & 0.983 & 0.974 & 0.949 & 0.920 \\
& \textit{Overall}
& 0.957 & 0.970 & 0.974 & 0.975 & 0.946 & 0.962 \\

\midrule

\multirow{4}{*}{Vegetation and natural damage}
& Low
& 0.966 & 0.861 & 0.991 & 0.962 & 0.974 & 0.898 \\
& Medium
& 0.932 & 0.936 & 0.923 & 0.881 & 0.915 & 0.898 \\
& High
& 0.949 & 0.944 & 0.932 & 0.924 & 0.949 & 0.942 \\
& \textit{Overall}
& 0.949 & 0.950 & 0.949 & 0.933 & 0.946 & 0.939 \\

\midrule

\multirow{4}{*}{Infrastructure and facility damage}
& Low
& 0.991 & 0.955 & 0.991 & 0.955 & 0.983 & 0.913 \\
& Medium
& 0.915 & 0.906 & 0.915 & 0.902 & 0.915 & 0.907 \\
& High
& 0.915 & 0.897 & 0.906 & 0.886 & 0.923 & 0.908 \\
& \textit{Overall}
& 0.940 & 0.944 & 0.937 & 0.941 & 0.940 & 0.946 \\

\midrule

\multirow{4}{*}{Debris and scene disruption}
& Low
& 0.966 & 0.895 & 0.949 & 0.855 & 0.949 & 0.855 \\
& Medium
& 0.923 & 0.918 & 0.906 & 0.846 & 0.915 & 0.861 \\
& High
& 0.957 & 0.941 & 0.957 & 0.941 & 0.932 & 0.903 \\
& \textit{Overall}
& 0.949 & 0.957 & 0.937 & 0.935 & 0.932 & 0.931 \\

\specialrule{0.9pt}{2pt}{2pt}
\multicolumn{2}{l!{\vrule width 0.6pt}}{\textbf{Overall indicator-level agreement}}
&  0.950 & 0.962  & 0.950 & 0.954 & 0.943 &  0.951 \\

\multicolumn{2}{l!{\vrule width 0.6pt}}{\textbf{Overall severity-level agreement}}
& 0.915 & 0.924 & 0.949 & 0.954 & 0.954 & 0.960 \\

\multicolumn{2}{l!{\vrule width 0.6pt}}{\textbf{Severity underestimation rate}}
& 0.066 & \textemdash & 0.040 & \textemdash & 0.014 & \textemdash \\

\multicolumn{2}{l!{\vrule width 0.6pt}}{\textbf{Severity overestimation rate}}
& 0.020 & \textemdash & 0.011 & \textemdash & 0.031 & \textemdash \\

\bottomrule
\end{tabular}
}
\end{table*}

The expert verification results in Table \ref{tab:vlm_indicator_agreement} show high agreement between the Gemini-generated annotations and the expert assessments. The verification protocol was relatively permissive: experts examined whether the reasoning scores were reasonable and supported by visible cross-view evidence, rather than requiring the generated explanations to reproduce expert interpretations exactly. Differences in wording or emphasis were accepted when they did not alter the main spatial judgment. Under this criterion, Gemini achieved overall indicator-level agreement values of 0.938 and 0.962 with Experts A and B, with corresponding quadratic weighted kappa values of 0.953 and 0.960. The agreement between the two experts was 0.934, with a $\kappa_w$ of 0.960. Although the evaluation protocol partly accounts for these high values, all expert-validation agreement and $\kappa_w$ results exceeded 0.8, indicating that the annotations were sufficiently reliable for model evaluation after expert quality control.

For the best-performing Task 1 results obtained using DINOv3-L, each VLM was required to determine whether the retrieved VGI--SVI--RSI triplet was correctly matched while performing spatial reasoning. GPT-4o achieved the highest matching accuracy at 69.00\%, followed by Qwen3.6-27B at 65.15\% and InternVL3-8B at 64.20\%. These results were obtained through zero-shot annotation without task-specific training and were numerically higher than those of the evaluated deterministic and probabilistic cross-view geolocalization methods. GPT-4o also achieved the strongest aggregate spatial reasoning performance, with an overall indicator-level agreement of 0.646 and a $\kappa_w$ of 0.489, compared with 0.609 and 0.396 for Qwen3.6-27B and 0.465 and 0.284 for InternVL3-8B. GPT-4o performed particularly well on building-related indicators, with agreement values ranging from 0.672 to 0.724. Its $\kappa_w$ reached 0.630 for recognizable building outlines, 0.613 for overall building appearance, and 0.610 for building type and scale. It also achieved agreement values of 0.751 and 0.731 for scene functional type and spatial density or openness. Its weaker results were concentrated in waterbody position, terrain outline, large facility structures, and regional background, where $\kappa_w$ was generally below 0.35.

Qwen3.6-27B performed relatively well on relational and scene-level indicators. Its highest agreement was obtained for VGI--SVI--RSI scene consistency at 0.721, followed by the scene functional type at 0.708, building--road--vegetation relationship at 0.700, facility grouping arrangement at 0.683, and spatial density and openness at 0.682. It also produced relatively high $\kappa_w$ values for traffic markers, utility poles and cables, fences and guardrails, and relationships among buildings, roads, and vegetation. Performance declined for coastline and terrain outline, waterbody and open-space position, and large facility structures. For large facility structures, an agreement of 0.589 was accompanied by a $\kappa_w$ of only 0.112, indicating limited discrimination among less frequent reasoning levels. InternVL3-8B produced lower results across most semantic groups. Its stronger indicators were recognizable building outlines and commercial signs, with agreement values of 0.556 and 0.531 and $\kappa_w$ values of 0.396 and 0.367. Its $\kappa_w$ values for vegetation coverage, waterbody position, terrain outline, scene functional type, and regional background ranged from 0.126 to 0.173. Overall VGI--SVI--RSI scene consistency reached an agreement of 0.407 and a $\kappa_w$ of 0.186, reflecting difficulty in integrating evidence across viewpoints and spatial scales.

Structural and relational cues generally provided the most reliable and consistent evidence for validating and understanding the cross-view geolocalization results. Building geometry, road organization, distinctive intersections, facility arrangements, and recognizable signs remain comparatively stable across ground, street, and aerial views, whereas vegetation, terrain, waterbodies, coastlines, and regional backgrounds are more sensitive to temporal change, viewpoint occlusion, disaster disturbance, and scale differences. GPT-4o showed the most balanced performance across appearance, spatial layout, and global-scene indicators, while Qwen3.6-27B remained competitive in interpreting relational and scene-level cues. Spatial reasoning therefore provides a useful complementary signal for geolocalization validation, although its reliability depends on the visibility, stability, and distinctiveness of the available cross-view evidence. As shown in Fig.~\ref{fig:resultsCVGL}, the models generally agree on structural and road cues, whereas vegetation, terrain, and global-scene reasoning produces more partial or incorrect results. Final geolocalization judgments vary across models, also reflecting differences in cross-view evidence integration.

\subsection{VLM-based damage reasoning and classification results}

\begin{table*}[htbp]
\centering
\caption{Agreement and quadratic weighted kappa evaluation of VLMs for disaster damage severity assessment using Gemini as a human annotation substitute. Bold the best-performing results in each category.}
\label{tab:combined_vlm_damage_comparison}
\footnotesize
\renewcommand{\arraystretch}{1.10}
\setlength{\tabcolsep}{6pt}

\begin{tabular}{
>{\arraybackslash}m{3.6cm}
>{\arraybackslash}m{1.0cm}
!{\vrule width 0.6pt}
>{\centering\arraybackslash}p{1.5cm}
>{\centering\arraybackslash}p{1.5cm}
!{\vrule width 0.6pt}
>{\centering\arraybackslash}p{1.4cm}
>{\centering\arraybackslash}p{1.4cm}
!{\vrule width 0.6pt}
>{\centering\arraybackslash}p{2.2cm}
>{\centering\arraybackslash}p{2cm}
}
\toprule

\textbf{Indicator}
& \textbf{Level}
& \textbf{Qwen Agr.}
& \textbf{Qwen $\kappa_w$}
& \textbf{GPT Agr.}
& \textbf{GPT $\kappa_w$}
& \textbf{InternVL Agr.}
& \textbf{InternVL $\kappa_w$} \\
\midrule

\multirow{4}{*}{Flooding and water impact}
& Low
& 0.661 & 0.601 & \textbf{0.717} & \textbf{0.667} & 0.394 & 0.306 \\

& Medium
& \textbf{0.789} & \textbf{0.851} & 0.772 & 0.844 & 0.583 & 0.462 \\

& High
& 0.802 & 0.842 & \textbf{0.877} & \textbf{0.915} & 0.627 & 0.427 \\

& \textit{Overall}
& 0.764 & 0.831 & \textbf{0.777} & \textbf{0.848} & 0.551 & 0.465 \\
\midrule

\multirow{4}{*}{Building and structural damage}
& Low
& 0.829 & 0.581 & \textbf{0.927} & \textbf{0.591} & 0.423 & 0.175 \\

& Medium
& 0.735 & 0.757 & \textbf{0.764} & \textbf{0.810} & 0.239 & 0.161 \\

& High
& \textbf{0.755} & \textbf{0.749} & 0.682 & 0.737 & 0.554 & 0.122 \\

& \textit{Overall}
& 0.760 & 0.788 & \textbf{0.790} & \textbf{0.817} & 0.341 & 0.262 \\
\midrule

\multirow{4}{*}{Road and access disruption}
& Low
& 0.588 & 0.400 & \textbf{0.712} & \textbf{0.620} & 0.299 & 0.219 \\

& Medium
& \textbf{0.691} & 0.584 & 0.575 & \textbf{0.606} & 0.539 & 0.227 \\

& High
& \textbf{0.767} & 0.499 & 0.742 & \textbf{0.589} & 0.717 & 0.050 \\

& \textit{Overall}
& \textbf{0.683} & 0.669 & 0.640 & \textbf{0.693} & 0.522 & 0.382 \\
\midrule

\multirow{4}{*}{Vegetation and natural damage}
& Low
& 0.748 & 0.597 & \textbf{0.880} & \textbf{0.700} & 0.459 & 0.235 \\

& Medium
& 0.662 & 0.692 & \textbf{0.751} & \textbf{0.779} & 0.301 & 0.278 \\

& High
& \textbf{0.738} & 0.660 & 0.711 & \textbf{0.703} & 0.475 & 0.296 \\

& \textit{Overall}
& 0.695 & 0.718 & \textbf{0.776} & \textbf{0.780} & 0.370 & 0.352 \\
\midrule

\multirow{4}{*}{Infrastructure and facility damage}
& Low
& 0.727 & \textbf{0.377} & \textbf{0.794} & 0.333 & 0.289 & 0.024 \\

& Medium
& 0.540 & 0.446 & \textbf{0.561} & \textbf{0.570} & 0.161 & 0.058 \\

& High
& \textbf{0.656} & 0.383 & 0.481 & \textbf{0.423} & 0.563 & 0.054 \\

& \textit{Overall}
& 0.603 & 0.582 & \textbf{0.605} & \textbf{0.603} & 0.267 & 0.169 \\
\midrule

\multirow{4}{*}{Debris and scene disruption}
& Low
& \textbf{0.583} & \textbf{0.449} & 0.556 & 0.386 & 0.333 & 0.160 \\

& Medium
& \textbf{0.633} & \textbf{0.532} & 0.512 & 0.406 & 0.289 & 0.075 \\

& High
& \textbf{0.845} & \textbf{0.640} & 0.541 & 0.160 & 0.720 & 0.124 \\

& \textit{Overall}
& \textbf{0.663} & \textbf{0.657} & 0.528 & 0.527 & 0.383 & 0.266 \\

\specialrule{0.9pt}{2pt}{2pt}

\multicolumn{2}{l!{\vrule width 0.6pt}}{\textbf{Overall indicator-level agreement}}
& \textbf{0.695} & 0.737 & 0.686 & \textbf{0.738} & 0.405 & 0.331 \\

\multicolumn{2}{l!{\vrule width 0.6pt}}{\textbf{Overall severity-level agreement}}
& 0.711 & 0.630 & \textbf{0.739} & \textbf{0.688} & 0.338 & 0.226 \\

\multicolumn{2}{l!{\vrule width 0.6pt}}{\textbf{Severity underestimation rate}}
& 0.066 & \textemdash & 0.155 & \textemdash & \textbf{0.041} & \textemdash \\

\multicolumn{2}{l!{\vrule width 0.6pt}}{\textbf{Severity overestimation rate}}
& 0.223 & \textemdash & \textbf{0.106} & \textemdash & 0.620 & \textemdash \\

\bottomrule
\end{tabular}
\end{table*}

\begin{figure*}
    \centering
    \includegraphics[width=\linewidth]{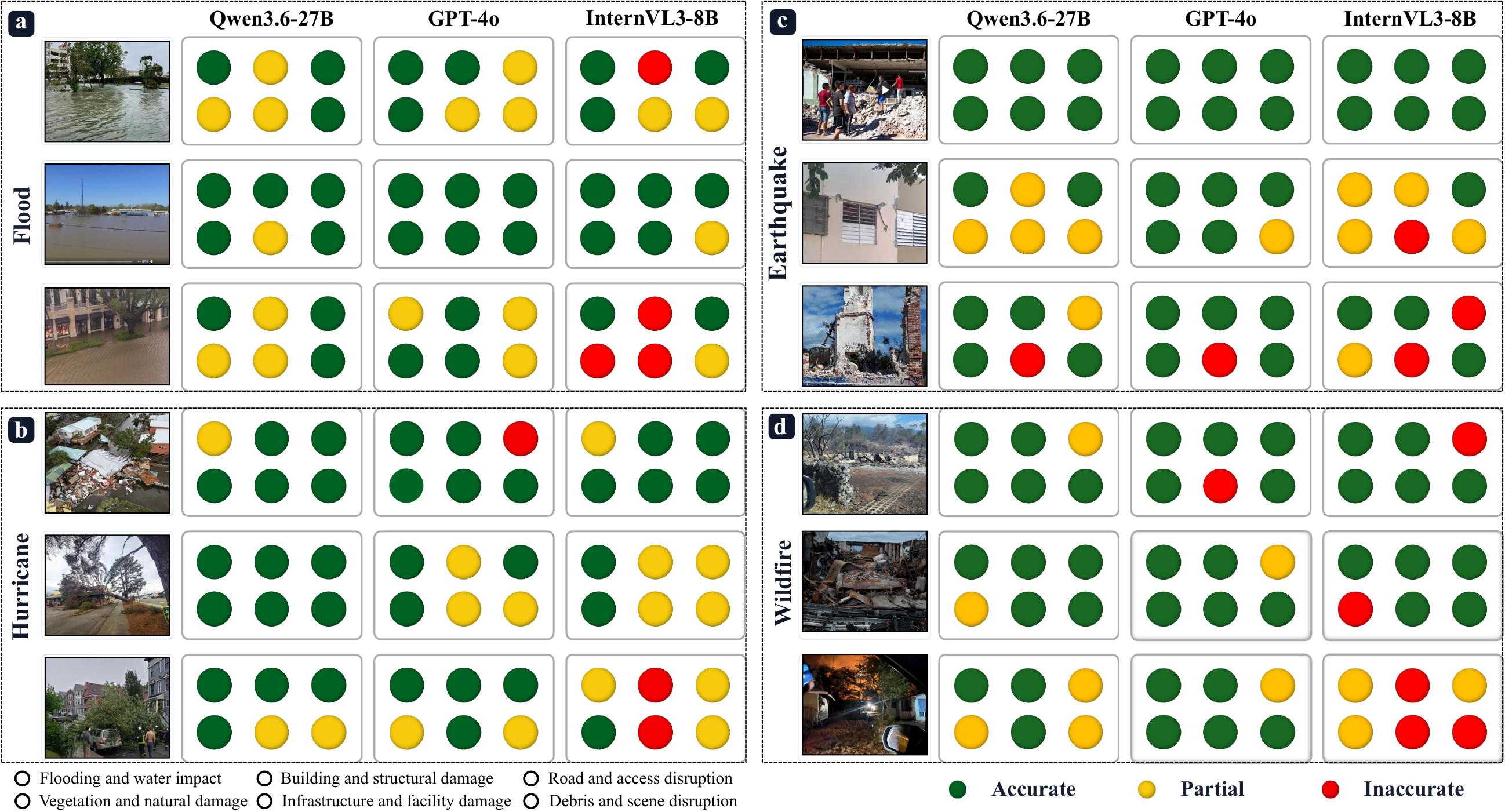}
    \caption{Examples of VLM-based disaster damage reasoning across six dimensions. Three VLMs are compared in flooding and water impact, building and structural damage, road and access disruption, vegetation and natural damage, infrastructure and facility damage, and debris and scene disruption. The circles in each group follow this order, with different colors indicating accuracy.}
    \label{fig:resultsDamage}
\end{figure*}

The expert verification results in Table~\ref{tab:expert_vlm_damage_comparison} support the use of Gemini-generated damage severity annotations. At the indicator level, Gemini achieved an agreement of 0.950 with both Experts A and B, with corresponding $\kappa_w$ values of 0.962 and 0.954. These values are comparable to the expert-to-expert agreement of 0.943 and $\kappa_w$ of 0.951. For overall severity, Gemini achieved agreement values of 0.915 and 0.949 with the two experts and $\kappa_w$ values of 0.924 and 0.954, respectively. The corresponding expert-to-expert values were 0.954 and 0.960. Gemini showed limited directional bias, with underestimation rates of 0.066 and 0.040 and overestimation rates of 0.020 and 0.011. Agreement across individual indicators ranged from 0.934 to 0.974. The consistency with both experts indicates that the generated annotations are sufficiently reliable for large-scale evaluation after expert quality control.

Table \ref{tab:combined_vlm_damage_comparison} compares Qwen3.6-27B, GPT-4o, and InternVL3-8B against the verified Gemini-based annotations. Qwen3.6-27B and GPT-4o produced similar aggregate indicator-level performance, although their strengths differed across damage types. Qwen3.6-27B obtained an overall indicator-level agreement of 0.695 and a $\kappa_w$ of 0.737, whereas GPT-4o achieved 0.686 and 0.738, respectively. GPT-4o performed better for overall severity classification, reaching an agreement of 0.739 and a $\kappa_w$ of 0.688, compared with 0.711 and 0.630 for Qwen3.6-27B. InternVL3-8B was substantially less consistent, with an indicator-level agreement of 0.405 and a $\kappa_w$ of 0.331, and an overall severity agreement of 0.338 with a $\kappa_w$ of 0.226. The gap suggests that reliable damage assessment requires visual object recognition and the integration of multiple damage cues into an ordered scene-level judgment.

At the category-overall level, GPT-4o achieved the strongest results for flooding and water impact, building and structural damage, vegetation and natural damage, and infrastructure and facility damage. Its overall agreement and $\kappa_w$ reached 0.777 and 0.848 for flooding, 0.790 and 0.817 for building damage, and 0.776 and 0.780 for vegetation damage. However, no model consistently performed best across all severity levels. For example, Qwen3.6-27B performed best for medium-level flooding, whereas GPT-4o performed best for low- and high-level flooding. Qwen3.6-27B also achieved the best results for high-level building damage, although GPT-4o performed better for this category overall. Results for road disruption and infrastructure damage were similarly mixed between agreement and $\kappa_w$. Qwen3.6-27B showed a clearer advantage for debris and scene disruption across all severity levels, obtaining an overall agreement of 0.663 and a $\kappa_w$ of 0.657, compared with 0.528 and 0.527 for GPT-4o. Infrastructure and facility damage remained difficult for both leading models, with overall agreement close to 0.60, possibly because utility failures, facility functionality, and partially occluded infrastructure are difficult to assess from ground-level disaster photographs.

The error distributions reveal different tendencies among the evaluated models. Qwen3.6-27B had a relatively low underestimation rate of 0.066 but an overestimation rate of 0.223, indicating a tendency to assign higher severity levels when errors occurred. GPT-4o showed a more balanced profile, with underestimation and overestimation rates of 0.155 and 0.106, respectively. InternVL3-8B produced an overestimation rate of 0.620 despite an underestimation rate of only 0.041, indicating frequent assignment of higher-than-reference severity levels and helping to explain its weak agreement and kappa scores. GPT-4o was the most reliable model for overall severity classification, while Qwen3.6-27B provided competitive indicator-level consistency and performed particularly well for debris-dominated scenes. The remaining errors mainly involved ambiguous cases with overlapping damage indicators, unclear functional impacts, or severity distinctions requiring evidence beyond the image.

The examples in Fig.~\ref{fig:resultsDamage} show how the three VLMs interpret the same damage evidence. In the flooded scene, all models identify severe water impact and restricted road access but differ in their assessments of buildings, vegetation, and infrastructure. In the structural-collapse scene, they consistently recognize collapsed walls and heavy debris while interpreting road accessibility and facility damage differently. The main damage type is generally recognized, whereas secondary indicators and functional impacts produce more disagreement.

\section{Discussion} \label{Discussions}

Cross-view geolocalization provides the spatial anchor for interpreting crowdsourced disaster imagery. Unlike conventional queries, disaster VGI may depict flooded roads, collapsed buildings, debris-covered surroundings, or disturbed vegetation that no longer resembles the available SVI and RSI references. Stronger foundation-model representations improve the extraction of persistent geographic features, while multi-view fusion reduces dependence on any single source. VGI retains local objects and visible damage, SVI provides street-level context and surrounding structures, and RSI represents broader layouts that may remain recognizable despite local disturbance. Retrieval-based methods are better suited to fine-grained geolocalization when the reference gallery has adequate coverage, whereas probabilistic models represent plausible geographic regions and uncertainty when the available cues are incomplete \citep{li2026towards, dufour2025around}. Disaster geolocalization can therefore benefit from combining precise reference matching with uncertainty-aware estimation rather than relying on a single geolocalization mechanism.

Retrieval scores cannot explain why a candidate geolocation is plausible or reveal whether high similarity reflects genuine geographic correspondence. This limitation becomes important when different places contain similar buildings, road patterns, vegetation, or urban forms. Although recent remote-sensing VLMs have demonstrated strong capabilities in grounded conversation and region-level image interpretation \citep{kuckreja2024geochat,liu2024remoteclip}, their potential for reasoning over cross-view geolocalization results remains insufficiently studied. GRDisaster applies VLM-based spatial reasoning to the retrieved VGI--SVI--RSI correspondences and organizes the supporting evidence into a dedicated set of cross-view reasoning indicators. Building geometry, road organization, intersections, facility arrangements, environmental features, and global-scene characteristics are evaluated separately rather than compressed into one similarity score. Structural and relational cues generally provide stronger evidence because their relative configurations can remain identifiable after substantial appearance changes, whereas vegetation, terrain, waterbodies, and regional backgrounds are more sensitive to season, scale, viewing direction, and disaster disturbance. The indicator structure makes disagreements traceable to specific cues and helps distinguish geographically meaningful correspondence from broad semantic resemblance.

The second set of reasoning indicators addresses disaster damage severity through flooding, structural damage, road access, vegetation disturbance, infrastructure condition, and debris. Existing disaster assessment studies have largely concentrated on change detection and building-level damage classification from pre- and post-event satellite imagery \citep{gupta2019creating}. GRDisaster instead extends VLM reasoning from determining whether images represent the same place to interpreting multiple forms of damage visible at that place. The models generally recognize dominant damage patterns, while greater disagreement appears in secondary effects and functional impacts. Standing water may be visible without establishing whether a road remains passable, and exterior damage may not reveal whether a building or facility remains operational. Such ambiguities explain why infrastructure functionality, accessibility, and adjacent severity levels are harder to assess than directly observable flooding or collapse. Differences in overestimation and underestimation also indicate that reliable severity assessment depends on calibration as well as visual recognition. Applying VLM-based reasoning to both geolocalization validation and damage assessment distinguishes GRDisaster from studies that address these tasks independently. Expert verification remains necessary when indicators conflict or judgments depend on evidence beyond the image. The two indicator systems should be regarded as an initial and extensible organization of relevant evidence rather than a complete or definitive taxonomy. Future studies may refine, prioritize, or expand them for different disasters, geographic contexts, and operational needs.

The two indicator systems connect geographic credibility with disaster impact while preserving the different roles of the underlying methods. Roads, buildings, waterbodies, vegetation, and facilities may support a geolocation match and also reveal how the event has altered access, structure, and environmental conditions. Deterministic retrieval supplies precise reference-based anchoring, probabilistic geolocalization describes spatial uncertainty, and VLM reasoning examines the evidence behind correspondence and severity judgments. Crowdsourced imagery can thus be interpreted as a georeferenced record of local conditions rather than as an isolated visual observation. The outputs should nevertheless be treated as structured decision support rather than autonomous conclusions, especially when reference imagery is outdated, gallery coverage is uneven, important objects are occluded, or the scene differs from the regions and disaster types represented in PhotoMappers. Single-image reasoning also has limited ability to determine hidden structural damage, infrastructure functionality, and the temporal progression of an event.

Considering model capability, availability, inference cost, and representation across different developers, the evaluation includes Qwen3.6-27B \citep{qwen2026qwen36}, GPT-4o \citep{openai2024gpt4o}, and InternVL3-8B \citep{zhu2025internvl3}. The findings provide evidence for the effectiveness of GRDisaster and VLM-based spatial reasoning, but they are limited to the evaluated models and versions rather than VLMs as a whole, particularly because commercial and open-source models evolve rapidly and may respond differently to prompt design. Future evaluation should cover a broader range of model families, parameter scales, and deployment settings, with explicit comparisons of reasoning accuracy, computational cost, latency, and prompt sensitivity. Pre- and post-event image sequences, road networks, elevation, weather records, and infrastructure databases could help separate persistent geographic structures from temporary disaster effects. Additional regions, event categories, image sources, reference densities, and temporal gaps are also needed to assess geographic transfer and disaster-type generalization. Calibrated indicator-level confidence, explicit recognition of insufficient visual evidence, lightweight models for field deployment, and selective expert review could further improve the reliability and practical value of VLM-assisted disaster mapping.

\section{Conclusions} \label{Conclusions}

This study introduced GRDisaster, a multi-task geospatial reasoning framework for converting crowdsourced disaster imagery into georeferenced and interpretable information. Built on the newly curated PhotoMappers dataset of 8,780 human-validated VGI--SVI--RSI cross-view triplets (26,340 images), it combines deterministic and probabilistic cross-view geolocalization with explainable VLM-based reasoning for match verification and damage severity classification. The results show that foundation-model features, complementary cross-view information, and optional re-ranking improve geographic matching. Two structured indicator systems support interpretable spatial validation and damage assessment, while expert verification confirms the reliability of the generated annotations. Tests with Qwen3.6-27B, GPT-4o, and InternVL3-8B demonstrate meaningful zero-shot reasoning, although performance varies across models and evidence types. GRDisaster provides a scalable and expert-verifiable framework for GeoAI-assisted disaster mapping, but its generalizability remains limited by reference coverage, limited visual evidence from individual images, and the range of evaluated disasters and models. Future work should evaluate broader regions, disaster types, and VLMs and incorporate temporal and auxiliary geospatial information to improve generalization and reliability.

{\small
\bibliography{references}
\bibliographystyle{IEEEtran}}

\vfill

\end{document}